\documentclass[lettersize,journal]{IEEEtran}
\usepackage{amsmath,amsfonts}
\usepackage{algorithmic}
\usepackage{algorithm}
\usepackage{array}
\usepackage[caption=false,font=normalsize,labelfont=sf,textfont=sf]{subfig}
\usepackage{textcomp}
\usepackage{stfloats}
\usepackage{url}
\usepackage{verbatim}
\usepackage{graphicx}
\usepackage{cite}
\usepackage{xcolor} 
\begin{document}

% -----------------------------------------------------------------------------------

% Improving facial attribute recognition by group and graph learning

% Neural Vector Fields: Implicit Representation by Explicit Learning

\title{An Efficient Adaptive Compression Method for Human Perception and Machine Vision Tasks}

\author{Lei Liu, Zhenghao Chen, Zhihao Hu, Dong Xu,~\IEEEmembership{Fellow,~IEEE}
\vspace{-8mm}
        % <-this % stops a space
\thanks{Lei Liu and Zhihao Hu are with the School of Computer Science and Engineering, Beihang University, Beijing 100000, China. E-mail: liulei95@buaa.edu.cn; huzhihao@buaa.edu.cn.}
\thanks{Zhenghao Chen is with the School of Information and Physical Sciences, University of Newcastle, New South Wales, NSW2300, Australia. E-mail: zhenghao.chen@newcastle.edu.au. }
\thanks{Dong Xu are with the Department of Computer Science, University of Hong Kong, Hong Kong SAR, China. E-mail: dongxu@hku.hk.}
\thanks{Corresponding author: Dong Xu.}}
% \thanks{This paper was produced by the IEEE Publication Technology Group. They are in Piscataway, NJ.}% <-this % stops a space
% \thanks{Manuscript received April 19, 2021; revised August 16, 2021.}}

% The paper headers
\markboth{Journal of \LaTeX\ Class Files,~Vol.~14, No.~8, August~2021}%
{Shell \MakeLowercase{\textit{et al.}}: A Sample Article Using IEEEtran.cls for IEEE Journals}

% \IEEEpubid{0000--0000/00\$00.00~\copyright~2021 IEEE}
% Remember, if you use this you must call \IEEEpubidadjcol in the second
% column for its text to clear the IEEEpubid mark.

\maketitle

\begin{abstract}

While most existing neural image compression (NIC) and neural video compression (NVC) methodologies have achieved remarkable success, their optimization is primarily focused on human visual perception. However, with the rapid development of artificial intelligence, many images and videos will be used for various machine vision tasks. Consequently, such existing compression methodologies cannot achieve competitive performance in machine vision. 
In this work, we introduce an efficient adaptive compression (EAC) method tailored for both human perception and multiple machine vision tasks. 
Our method involves two key modules: 
1), an adaptive compression mechanism, that adaptively selects several subsets from latent features to balance the optimizations for multiple machine vision tasks (\textit{e.g.}, segmentation, and detection) and human vision.
2), a task-specific adapter, that uses the parameter-efficient delta-tuning strategy to stimulate the comprehensive downstream analytical networks for specific machine vision tasks.
By using the above two modules, we can optimize the bit-rate costs and improve machine vision performance.
In general, our proposed EAC can seamlessly integrate with existing NIC (\textit{i.e.}, Ballé2018, and Cheng2020) and NVC (\textit{i.e.}, DVC, and FVC) methods. Extensive evaluation on various benchmark datasets (\textit{i.e.}, VOC2007, ILSVRC2012, VOC2012, COCO, UCF101, and DAVIS) shows that our method enhances performance for multiple machine vision tasks while maintaining the quality of human vision.

 % Recent coding for machine methods are typically designed with very specific codecs or vision networks, which limits the methods' universality across various codecs and visual tasks, and may contribute to a significant amount of computational consumption.
 % %
 % This study introduces an efficient adaptive compression (EAC) method, which is tailored for human perception and multiple machine vision tasks. 
 % % xxx
 % This method involves two key strategies: First, we introduce a novel approach to adaptively compress the latent representation of input data for multiple machine vision tasks and human vision. Second, to effectively optimize the machine vision tasks, we employ a parameter-efficient delta-tuning approach by utilizing a task-specific adapter that stimulates a large-parameter network, fine-tuned for particular machine vision tasks. Our proposed EAC method embodies universality, and it can seamlessly integrate with existing NIC (\textit{e.g.}, Ballé2018, Cheng2020) and NVC (\textit{e.g.}, DVC, and FVC) methods. We have conducted extensive testing on various benchmark datasets (\textit{i.e.}, VOC2007, ILSVRC2012, VOC2012, COCO, UCF101, and DAVIS), and our results demonstrate enhanced performance for multiple machine vision tasks, all while maintaining the quality of human vision.
\end{abstract}


\section{Introduction}
% importance
\IEEEPARstart{A}{S} machine learning for multimedia analytical tasks is rapidly evolving, bringing forth new challenges in efficiently storing image and video data for machine vision. This issue is particularly critical as most existing compression methods are designed primarily for human visual perception. Therefore, developing a compression method tailored specifically for machine vision has become imperative.
Recently, there have been several attempts to use neural networks for image and video compression tailored to machine vision~\cite{liu2021semantics, torfason2018towards, song2021variable, wang2021end, mei2021learn, 9414465,chen2023transtic}. However, these methods typically incorporate optimization terms for specific vision tasks, which only compromises human visual performance. 

Recently, another category of methods~\cite{choi2022scalable, bai2022towards, 9385898, wang2021towards,lin2023deepsvc} has introduced scalable strategies to balance optimization for both human and machine perception. 
Despite their promise, these scalable methods have three significant limitations.
First, the codecs in these methods are merely optimized for a single machine vision task, making it difficult to handle multiple machine vision tasks simultaneously in practical applications. For example, Bai \textit{et al.}~\cite{bai2022towards} designed an image compression method solely for the classification task.
% First, they are only optimized for a single machine vision task, without considering their simultaneous use for multiple vision tasks (\textit{e.g.}, segmentation, and detection). 
%
Second, they usually require an end-to-end optimization for task-specific networks for machine vision tasks, which becomes unscalable as the networks grow more comprehensive and compass more parameters.
%
% Third, 去调研一下是不是你是第一个可以同时做image和video的，是的说别人都不行
Third, to the best of our knowledge, most methods design a specific codec only for either images~\cite{liu2021semantics, torfason2018towards, song2021variable, wang2021end, mei2021learn, 9414465,choi2022scalable, bai2022towards, 9385898, wang2021towards,chen2023transtic} or videos~\cite{lin2023deepsvc,tian2023non}, resulting in a lack of generalization for processing both images and videos.

% adaptive, multi—task
% 重写
% 总分总
% 先说你propose了一个XXX，然后那些模块分别解决了上面三个问题

To tackle the above-mentioned limitations, we propose an Efficient Adaptive Compression (EAC) method for human perception and multiple vision tasks, which contains an adaptive compression module and task-specific adapters. More specifically, 
to simultaneously handle the multiple vision tasks, we have introduced an \textbf{adaptive} compression method, which allows us to partition, transmit, reconstruct, and aggregate the quantized latent representation in an adaptive manner for multiple machine vision tasks and human vision. Specifically, our approach partitions the quantized latent representation into several subsets for different machine vision tasks by using some binary masks which are generated by a hyperprior network and some predictors. Our method adaptively transmits different subsets for different machine vision tasks, which helps save bit-rate costs. Only when a high-quality data reconstruction is required for human vision, the full latent representation will be transmitted.
%
% This adaptive compression method will save bit-rate costs for machine vision tasks and will reconstruct high-quality data for human vision.

% efficient 现状+challenge+你做了什么+效果
To enhance the efficiency of optimization in our human-machine-vision compression methods,
drawing inspiration from the parameter-\textbf{efficient} delta-tuning~\cite{ding2023parameter} strategy used in large foundation models, we propose the integration of task-specific adapters. Each adapter is designed to stimulate the task-specific network for downstream machine vision tasks. Crucially, we optimize only this adapter, which has a relatively small number of parameters, while keeping the comprehensive task-specific network frozen. 
This approach allows us to achieve efficient optimization with significantly reduced computational costs.

% 重写，这是你的novelty你要说你的方法可以同时用在image和video上，说image和video的设计有什么不一样，video要利用temporal信息，所以你用了reference信息， 另外resp不是这么用的，分开写，先说image，后说video
% used in image compression and video compression
% By incorporating the aforementioned adaptive compression method and efficient optimization strategy, we introduce an Efficient Adaptive Compression (EAC) Method for human perception and multiple vision tasks. 
% Our EAC is generalization for processing both images and videos, which is readily applied EAC to various neural image compression methods (\textit{i.e.}, Ballé2018~\cite{ballevariational} and Cheng2020~\cite{cheng2020learned}) and various neural video compression methods (\textit{i.e.}, DVC~\cite{dvc} and FVC~\cite{fvc}).
Our EAC shows generalization in processing both images and videos, being readily applicable to various neural image compression methods (\textit{i.e.}, Ballé2018~\cite{ballevariational} and Cheng2020~\cite{cheng2020learned}) and neural video compression methods (\textit{i.e.}, DVC~\cite{dvc} and FVC~\cite{fvc}).
For image compression, with the help of spatial information (\textit{i.e.}, the distribution of the latent representations or the reconstructed image), we adopt the adaptive compression module for the latent representations of images to save bit-rate costs and use the task-specific adapter to optimize the task-specific network effectively. For video compression, both spatial information (\textit{i.e.}, the distribution of motion and residual features or the reconstructed frame) and temporal information (\textit{i.e.}, reference frames) are adopted by the adaptive compression module and task-specific adapter for better performance.

% We have conducted comprehensive experiments on various machine vision benchmarks, including ILSVRC2012~\cite{deng2009imagenet}, VOC2007~\cite{voc}, VOC2012~\cite{voc}, COCO~\cite{coco}, UCF101~\cite{soomro2012ucf101}, and DAVIS~\cite{perazzi2016davis}, to demonstrate the superiority of our EAC method. The results reveal that our EAC method not only achieves significant bit-rate savings for multiple machine vision tasks (\textit{e.g.}, segmentation, detection), but also maintains excellent performance in human vision. Additionally, even though using codecs from such classic neural compression methods (\textit{i.e.}, Ballé2018~\cite{ballevariational}, Cheng2020~\cite{cheng2020learned}, DVC~\cite{dvc}, and FVC~\cite{fvc}), our method outperforms the latest standard VTM~\cite{VVC} in most bit-rates.
% The main contributions of this work can be summarized as follows:
We have conducted comprehensive experiments on various machine vision benchmarks, including ILSVRC2012~\cite{deng2009imagenet}, VOC2007~\cite{voc}, VOC2012~\cite{voc}, COCO~\cite{coco}, UCF101~\cite{soomro2012ucf101}, and DAVIS~\cite{perazzi2016davis}, to demonstrate the superiority of our EAC method. The results reveal that our EAC method not only achieves significant bit-rate savings for multiple machine vision tasks (\textit{e.g.}, segmentation, detection) but also maintains excellent performance in human vision. Moreover, even when using codecs from classic neural compression methods (i.e., Ballé2018~\cite{ballevariational}, Cheng2020~\cite{cheng2020learned}, DVC~\cite{dvc}, and FVC~\cite{fvc}), our method outperforms the latest standard VTM~\cite{VVC} across most bit-rates.

\begin{itemize}
    % 重写，highlight可以做多个任务
    \item We propose an adaptive compression method to simultaneously balance the optimizations for multiple machine vision tasks and human vision by adaptively selecting optimal subsets of quantized latent features.

    \item We utilize a parameter-efficient delta-tuning strategy through task-specific adapters, seamlessly integrated into our method. This approach offers a highly effective and lightweight alternative to the traditional practice of fine-tuning the entire task-specific network.

    \item By combining the adaptive compression module with the task-specific adapters, we introduce a general method called Efficient Adaptive Compression (EAC). Our EAC can be seamlessly integrated into various neural image compression methods (\textit{i.e.}, Ballé2018 and Cheng2020) and neural video compression methods (\textit{i.e.}, DVC and FVC), extending these methods to enhance machine perception.

    \item Extensive experimental results underscore the efficacy of our newly proposed EAC method. It excels in delivering promising results across multiple machine vision tasks while concurrently preserving the integrity of human visual performance.
    
\end{itemize}

% 重写，每个extension简单描述一下就好，然后你很多缩写前面解释了，不要比如 Neural Image Compressions (NICs)，为什么还要在解释一遍？能不能认真检查一下？
This work extends our preliminary conference version~\cite{liu2023icmh} with the following substantial improvements. 
1) We extend our conference-approved adaptive compression method to support multiple machine vision tasks by adaptively selecting different subsets of latent representation for different tasks. 
2) The adaptive compression method can be applied to various NVC methods. Our method uses both spatial information and temporal information to eliminate more redundancies in both motion and residual for successive machine vision tasks.
3) We utilize a parameter-efficient delta-tuning strategy by designing a new task-specific adapter. 
4) We provide more experiments and analysis to further verify the effectiveness of our EAC in different compression methods (\textit{i.e.}, Ballé2018~\cite{ballevariational}, Cheng2020~\cite{cheng2020learned}, DVC~\cite{dvc}, and FVC~\cite{fvc}) for both multiple machine vision tasks and human vision. In addition, we also perform more ablation studies and detailed analysis of our adaptive compression method and task-specific adapter including the effectiveness and complexity of different components, visualization, and so on.

\section{Related Work}

\subsection{Neural Image and Video Compression}
NIC methods~\cite{balle2017end,ballevariational,minnen2018joint,cheng2020learned,1,2,3,7,8,9,10,11,12,he2022elic,han2024cra5,liu2023icme} and NVC methods~\cite{lin2020mlvc,agustsson2020scale,djelouah2019neural,habibian2019video,hu2020improving,chen2022exploiting,hu2022coarse,lu2020content,lu2020end,yang2020learning,chen2021improving,liu2024towards,li2021deep,chen2022lsvc,sheng2022temporal,li2022hybrid,chen2023neural,chen2024group} have recently achieved promising compression performance. 
As an instance, Ballé \textit{et al.}~\cite{ballevariational} proposed an end-to-end optimized NIC by proposing the learned hyperprior network to capture spatial dependencies in the latent representation, later Cheng \textit{et al.}~\cite{cheng2020learned} proposed a simplified attention module and utilized the Gaussian Mixture Model to enhance the image compression network. 
DVC~\cite{dvc}, the first end-to-end optimized NVC, replaced all modules with neural networks, and FVC~\cite{fvc} later conducted all key coding operations within the feature space. 
However, most existing NIC and NVC methods are exclusively optimized for human vision, overlooking potential applications in machine vision.
 
Upon the success of such NIC methods and NVC methods, our method caters to both machine and human vision with an effective balance and achieves promising results for both machine vision and human vision, venturing beyond the capabilities of the previously discussed methods.

% \subsection{Coding for Both Machine and Human Vision}
\subsection{Compression for Machine Vision}
Machine intelligence advancements have highlighted the need for compressing large volumes of visual data crucial for machine vision. Recent studies have made considerable progress in NIC methods, benefitting both machine and human vision. 
For instance,~\cite{liu2021semantics} conceived a scalable image compression framework that allows for coarse-to-fine classification tasks,~\cite{bai2022towards} adopted the transformer-based method for both image compression classification tasks and human vision. Recently~\cite{liu2023icmh} splited the encoded and quantized feature into segments that can be independently transmitted for various machine vision tasks.
However, these methods have split the subset of the latent representation for one signal machine vision task only, which cannot be directly employed for multiple machine vision tasks. 
% we adopt xxx, xxxx
Additionally, these methods have been tailored predominately to reduce domain redundancy only, which cannot be directly used for video compression as they do not consider temporal redundancy.

Recently, the DeepSVC~\cite{lin2023deepsvc} dynamically segments quantized features for both human and machine vision branches. 
While its selection mechanism is overly complex and includes many unnecessary blocks that hinder effective optimization, our model takes a different approach by employing a lightweight network for adaptive selection, simplifying the optimization process.

Another common concern of existing human-vision-coding methods~\cite{liu2021semantics,bai2022towards,duan2020video} is that they usually need to optimize the whole task-specific network with the codec to achieve better machine vision performance. However, this makes optimization harder with the introduction of more complex task-specific networks. 
In our work, we mitigate this issue by adopting the efficient optimization strategy with a task-specific adapter with much less parameters. 

\subsection{Adapter-based Parameter-Efficient Delta-Tuning}
Adapters~\cite{ding2023parameter} serve as a category of practical and parameter-efficient delta-tuning methods. These approaches are known for their capability to seamlessly incorporate compact, fully interconnected networks directly into a pretrained model. The idea has gained significant attention~\cite{pfeiffer2020adapterhub}, leading to numerous modifications of adapters being made. The variations involve strategic changes in adapter positions~\cite{he2021towards,zhu2021counter}, the implementation of directed pruning techniques~\cite{he2022sparseadapter}, or the use of reparametrization methods~\cite{karimi2021compacter,edalati2022krona}.

In our proposed EAC, we introduce a lightweight adapter into the task-specific network for various downstream tasks. This adapter integrates information from the reconstructed image in image compression methods or from the current frame and multiple reference frames in video compression methods. By leveraging this adapter, our framework achieves a balance between performance on the downstream task and computational efficiency during fine-tuning.

\begin{figure*}[t!]
    \centering
    \includegraphics[width=\linewidth]{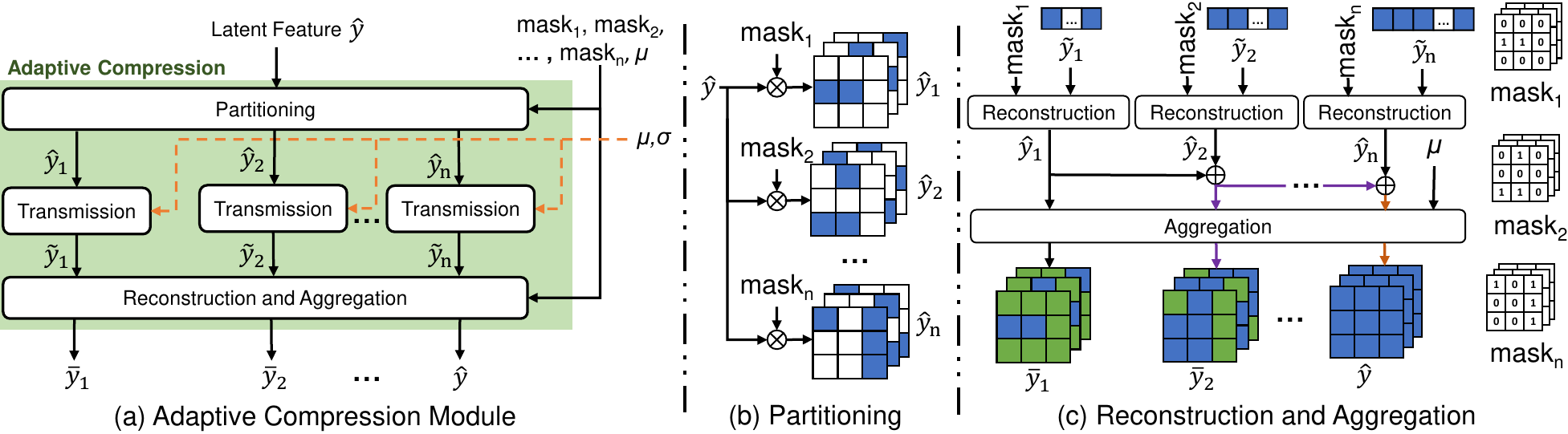}
    \vspace{-7mm}
    \caption{(a) The overview of our adaptive compression module, where we simultaneously balance the optimizations for the multiple machine and human vision tasks. The transmission module contains the arithmetic encoder and arithmetic decoder. (b) Details of the partitioning, which selects subsets from the quantized latent feature $\hat{y}$ for the various vision tasks. (c) Details of the reconstruction and aggregation modules. The reconstruction module reconstructs the quantized latent feature from 1D shape $\tilde{y}_i$ to 3D shape $\hat{y}_i$ using a predicted binary mask. The aggregation module fills unselected elements with their predicted mean value (\textit{i.e.}, $\mu$) for latent features.}
    \label{fig:adaptive_compression}
    \vspace{-4mm}
\end{figure*}

\section{Methodology}

Our Efficient Adaptive Compression (EAC) method comprises two key modules: the adaptive compression module and the task-specific adapter. In Section~\ref{section:split}, we introduce the adaptive compression module, which adaptively compresses images or videos for both multiple machine vision tasks and human vision. Next, in Section~\ref{section:adapter}, we describe the task-specific adapter, which utilizes the parameter-efficient delta-tuning strategy to stimulate the task-specific networks. The following Section~\ref{section:nic} and Section~\ref{section:nvc} detail how to apply our EAC to NIC methods and NVC methods, respectively. Finally, in Section~\ref{section:optimization}, we describe the optimization manner for the adaptive compression module and task-specific adapters.

\subsection{Adaptive Compression}
\label{section:split}
Considering the currently proposed methods in our conference version and other related works~\cite{liu2021semantics, torfason2018towards, song2021variable, wang2021end, mei2021learn, 9414465, liu2023icme,choi2022scalable, bai2022towards, 9385898, wang2021towards,liu2024towards,lin2023deepsvc} are only applicable for a single machine vision task, they fail to meet the requirements of multiple machine vision tasks (\textit{e.g.}, segmentation and detection) in complex scenarios (\textit{e.g.}, autonomous drive). 
To tackle this limitation, we designed the adaptive compression module by partitioning latent representation $\hat{y}$ into several subsets and adopting different subsets for different tasks to save bit-rate, as shown in Fig.~\ref{fig:adaptive_compression} (a).

Our adaptive compression module primarily processes the quantized latent feature $\hat{y}$, which is encoded and quantized from the uncompressed image in NIC methods or uncompressed motion/residual in NVC methods.
To select subsets of the quantized latent feature $\hat{y}$ for different tasks, the adaptive compression method partitions $\hat{y}$ into $n$ subsets (\textit{i.e.}, $\hat{y}_1$, $\hat{y}_2$, ..., and $\hat{y}_n$) through element-wise multiplication of $\hat{y}$ by binary masks (\textit{i.e.}, $\rm mask_1$, $\rm mask_2$, ..., $\rm mask_n$), as shown in Fig.~\ref{fig:adaptive_compression} (b). Each mask is produced by a predictor, designed identically to our conference version. 
Then the transmission module losslessly encodes the subset $\hat{y}_1$ into a bit-stream using an arithmetic encoder, which is then losslessly decoded to $\tilde{y}_1$ by an arithmetic decoder. Other subsets $\hat{y}_2, ..., \hat{y}_n$ are converted to $\tilde{y}_2, ..., \tilde{y}_n$ using the same transmission module, respectively.
To reduce bit-rate costs during transmission, we use a hyperprior network to produce hyperprior information (\textit{i.e.}, $\mu$ and $\sigma$) as in~\cite{ballevariational,cheng2020learned,liu2023icmh} to estimate the distribution of all subsets.

Note that the transmission module processes the latent representations to 1D shape (we denote them as $\tilde{y}_1, \tilde{y}_2, ..., \tilde{y}_n$), which are required to reshape into a 3D shape ($\hat{y}_1, \hat{y}_2, ..., \hat{y}_n$) for subsequent processing (\textit{e.g.}, 2D convolutional processing.). Hence, we adopt the reconstruction operation as in~\cite{lee2022selective} to convert the elements in a 1D shape $\tilde{y}_1, \tilde{y}_2, ..., \tilde{y}_n$ into elements of a 3D-shaped representation $\hat{y}_1, \hat{y}_2, ..., \hat{y}_n$ using the flattened index of ${\rm mask_1}, {\rm mask_2}, ..., {\rm mask_n}$, respectively.
After reconstruction, we aggregate 3D-shaped reconstructed latent representations for different tasks. Specifically, for the $i$-th machine vision task, we add reconstructed latent representations $\hat{y}_1$, ..., $\hat{y}_i$ in an element-wise way and fill the unselected elements with their predicted mean value (\textit{i.e.}, $\mu$) in the aggregation module to obtain the aggregated latent feature $\overline{y}_i$, where $\mu$ produces from the hyperprior network. Finally, the aggregated latent feature $\overline{y}_i$ is decoded to the image or video, which is then used by the subsequence task-specific network for machine vision.
For human vision, we add all reconstructed latent features $\hat{y}_1$, ..., $\hat{y}_n$ together in an element-wise way, obtaining full latent feature $\hat{y}$ to reconstruct the image or video in high quality.

Overall, in our adaptive compression module, the partitioning operation saves bit-rate costs for multiple machine vision tasks by adaptively selecting the different subsets from the full quantized latent representation $\hat{y}$, and the aggregation operation replaces the unselected elements with their predicted mean values for latent features to improve multiple machine vision performances without excess bit-rate costs.
 
\begin{figure}[!t]%[htbp]
\begin{center}
\includegraphics[width=\linewidth]{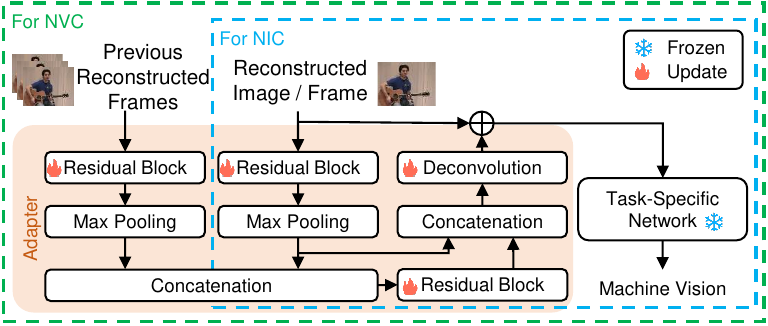}
\end{center}
\vspace{-5mm}
  \caption{The details of how we implement the adapter with the task-specific network.  We optimize it by using a parameter-efficient delta-tuning strategy. For NIC, our adapter only adopts spatial information (\textit{i.e.}, reconstructed image) as input. For NVC, our adapter utilizes both spatial information (\textit{i.e.}, current reconstructed frame) and temporal information (\textit{i.e.}, multiple previous reconstructed frames).}
  \vspace{-4mm}
\label{fig:adapter}
\end{figure}

% \subsection{Adapter-based Parameter-Efficient Delta-Tuning}
\subsection{Task-specific Adapter}
\label{section:adapter}
The previous NIC and NVC methods approach machine tasks by finetuning the task-specific network, which incurs high costs for complex tasks such as classification, segmentation, and detection. 
As an alternative to the full fine-tuning strategy, we introduce a lightweight task-specific adapter, designed to stimulate the following task-specific network inspired by recent parameter-efficient delta-tuning approach~\cite{ding2023parameter}. 
In sharp contrast with the previous approaches, the task-specific adapter achieves a good performance-cost trade-off by training only a few parameters with the task-specific network fixed. For instance, in the image classification task, the parameter number of the adapter is only 0.67\% compared with the task-specific network (\textit{i.e.}, ResNet50~\cite{resnet}).

As shown in Fig.~\ref{fig:adapter}. Firstly, we project the reconstructed image $\hat{X}_i$ (\textit{resp.}, reconstructed video frame $\hat{X}_{t}^{mi}$ and reference frames $\hat{X}_{t-1}^{mi}, \hat{X}_{t-2}^{mi}, \hat{X}_{t-3}^{mi}$) into the feature maps for image (\textit{resp.}, video) compression. Then, the previous features are combined into a new feature map through a single residual block. Finally, the intermediate features are concatenated and projected to the residual through the deconvolutional layers, yielding a new image or frame. This new image or frame is then fed into the task-specific network for downstream machine vision tasks. For image compression, the above process is formulated as:
\vspace{-2mm}
\begin{equation}
\begin{aligned}
\label{eq:adapter_image}
    P_i = f_\phi(\hat{X}_i + f_\psi(\hat{X}_i)),
\end{aligned}
\vspace{-2mm}
\end{equation}
where $P_i$ is the prediction for the $i$-th downstream task (\textit{e.g.}, segmentation map for segmentation task), $f_\phi$ is the task-specific network, $f_\psi$ is the adapter. For video compression, the adapter further adopts the temporal information from previous reconstructed frames (\textit{i.e.}, $\hat{X}_{t-1}^{mi}, \hat{X}_{t-2}^{mi}, \hat{X}_{t-3}^{mi}$) than it utilized in NIC methods. The process for video compression method is formulated as:
\begin{equation}
\begin{aligned}
\vspace{-1mm}
\label{eq:adapter_video}
    P_i = f_{\phi}(\hat{X}_t^{mi}\!+\!f_{\psi}(\hat{X}_t^{mi}, [\hat{X}_{t-1}^{mi}, \hat{X}_{t-2}^{mi}, \hat{X}_{t-3}^{mi}])),
\end{aligned}
\vspace{-1mm}
\end{equation}
where $[\cdot]$ is the concatenation operation.

\begin{figure}[t!]
    \centering
    \includegraphics[width=\linewidth]{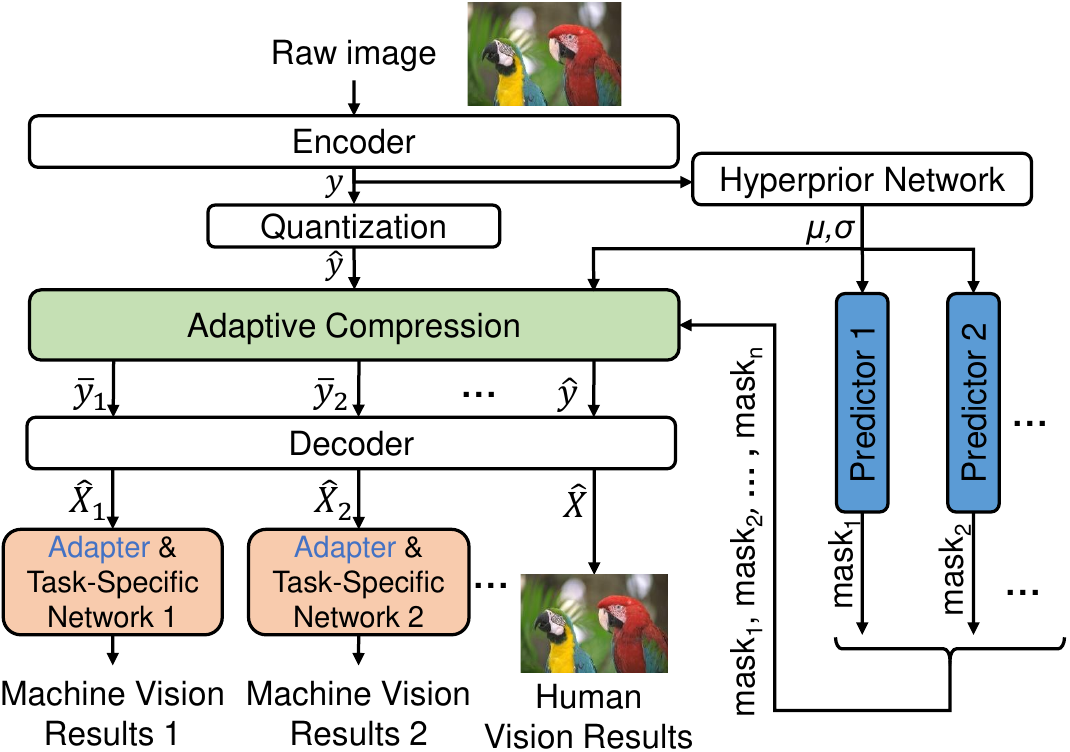}
    \vspace{-7mm}
    \caption{The overview of ``EAC (NIC)'', where we incorporate our efficient adaptive compression method in neural image compression network. }
    \label{fig:overview_image}
    \vspace{-5mm}
\end{figure}

\begin{figure*}[htbp]
    \centering
    \includegraphics[width=\textwidth]{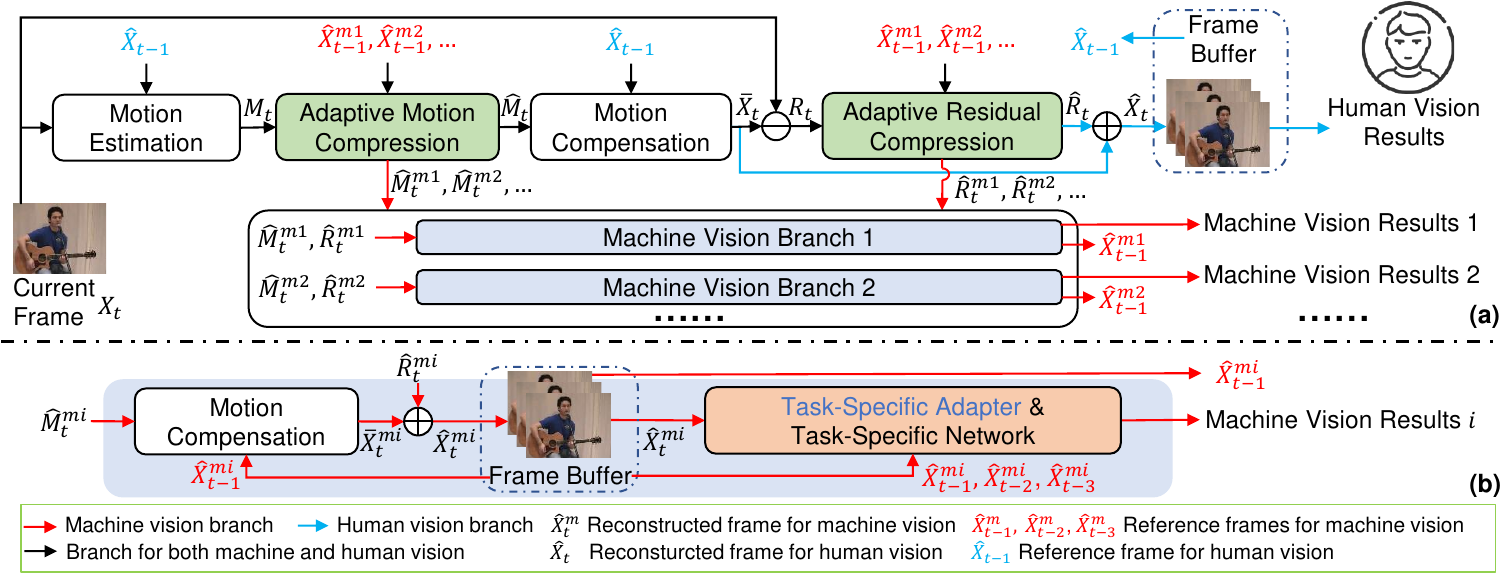}
    \vspace{-7mm}
  \caption{(a) The overview of ``EAC (NVC)'', where we incorporate our efficient adaptive compression method in neural video compression network. (b) The details of $i$-th machine vision branch, $i\in\{1,2,...,n\}$. Given an input frame $X_t$ at current time-step $t$, we first estimate the motion $M_t$ between $X_t$ and reference frame $\hat{X}_{t-1}$, which is then compressed by an adaptive motion compression network to produce the reconstructed motions $\hat{M}_t^{mi}, i \in \{1,2,...,n\}$ for machine vision (\textit{resp.}, $\hat{M}_t$ for human vision). Then we will adopt the reconstructed motion to perform motion compensation and predict frame $\overline{X}_t^{mi}$ for machine (\textit{resp.}, $\overline{X}_t$ for human). Then we can compress the residual $R_t$, which is produced by subtracting predicted frame $\overline{X}_t$ from the current frame $X_t$, by using adaptive residual compression network to produce the residual $\hat{R}_t^{mi}$ for machine vision (\textit{resp.}, $\hat{R}_t$ for human vision). Last, the predicted frame $\overline{X}_t^{mi}$ (\textit{resp.}, $\overline{X}_t$) will be then added back to reconstructed residual $\hat{R}_t^{mi}$ (\textit{resp.}, $\hat{R}_t$) to generate the final reconstructed frame $\hat{X}_t^{mi}$ for machine vision (\textit{resp.}, $\hat{X}_t$ for human vision).}
    \label{fig:overview_video}
    \vspace{-5mm}
\end{figure*}

\subsection{EAC Integration in NIC}
\label{section:nic}
We integrate our EAC method into the NIC framework (\textit{e.g.}, Ballé2018~\cite{ballevariational}), namely ``EAC (NIC)'', as shown in Fig.~\ref{fig:overview_image}.
``EAC (NIC)'' utilizes raw images as input, compressing them for machine or human vision. For example, during autonomous driving, ``EAC (NIC)'' is mostly used for multiple machine vision tasks by selectively transferring several subsets of the full latent representation to save bit-rate costs. When a traffic accident happens, ``EAC (NIC)'' will work for human vision by transferring the full latent representation which can reconstruct images with high visual quality for further analysis. The procedure of ``EAC (NIC)'' is summarized as follows:

The input image is transformed into a latent representation $y$ by the encoder, which is then quantized to $\hat{y}$ by the quantization module. 
Meanwhile, latent representation $y$ is utilized by a hyperprior network to produce the hyperperior information (\textit{i.e.}, $\mu$ and $\sigma$), which is then adopted by all predictors to generate binary masks for adaptive compression module. Specifically, predictor 1 takes $\mu$ and $\sigma$ to generate $\rm mask_1$ by using three convolutional layers, two leaky ReLU layers, and one Gumbel-Softmax/Max module, as described in our conference version. Other predictors (\textit{i.e.}, Preditor $i$, $i\in\{2, ...,n\}$) follow the same operation to generate binary masks (\textit{i.e.}, $\rm mask_i$), respectively.
With the help of the hyperprior information and all binary masks, our adaptive compression module adaptively compresses the quantized latent representation $\hat{y}$ into $\Tilde{y}_1$, $\Tilde{y}_2$, ..., $\Tilde{y}_n$, as shown in Fig.~\ref{fig:adaptive_compression}. 
Next, each reconstructed latent feature (\textit{e.g.}, $\Tilde{y}_1$) is reconstructed into image by the decoder to serve human vision perception or the subsequent multiple task-specific networks (\textit{i.e.}, ResNet50~\cite{resnet}, PSANet~\cite{zhao2018psanet}, and Faster R-CNN~\cite{frcnn}). 

Different from NIC methods that directly compress the whole quantized latent feature $\hat{y}$, our ``EAC (NIC)'' adaptively compresses $\hat{y}$ to several subsets for multiple machine vision tasks and human vision using our adaptive compression module described in Section~\ref{section:split}. 
Furthermore, to more effectively optimize task-specific networks, ``EAC (NIC)'' adopts a parameter-efficient delta-tuning approach by introducing task-specific adapters (described in Section~\ref{section:adapter}) with far fewer parameters than the original task-specific networks. By using the adaptive compression module and task-specific adapters, our ``EAC (NIC)'' can efficiently and adaptively compress the images for multiple machine vision tasks and human vision.

\subsection{EAC Integration in NVC}
\label{section:nvc}

To further demonstrate the generalization of our newly proposed EAC method, we incorporate it in the NVC network, namely ``EAC (NVC)''. 
Let $\{X_1, ..., X_{t}, ...\}$ denote a video sequence, where $X_t$ is the original video frame at the current time step $t$. 
As shown in Fig.~\ref{fig:overview_video}, ``EAC (NVC)'' compresses the video sequence for multiple machine vision tasks or human vision, producing compressed sequences $\{\hat{X}_1^{m1}, ..., \hat{X}_t^{m1}, ...\}, \{\hat{X}_1^{m2}, ..., \hat{X}_t^{m2}, ...\},...$ or $\{\hat{X}_1, ..., \hat{X}_{t}, ...\}$, respectively.
Similar to the ``EAC (NIC)'' method, ``EAC (NVC)'' is primarily used for machine vision. When human intervention is necessary, ``EAC (NVC)'' will work for humans.
The procedure of ``EAC (NVC)'' is described as follows:

The input current frame $X_t$ and the previous reconstructed frame $\hat{X}_{t-1}$ are converted to the motion $M_t$ by a motion estimation module, which is then compressed to $\mathrm{\hat{y}_t^{moti}}, i\in\{2,3,...\}$ and $\mathrm{\hat{y}_t^{mot}}$ by our adaptive motion compression module. Particularly in $i$-th machine vision branch (\textit{resp.}, human vision branch), given the reconstructed motion $\hat{M}_t^{mi}$ (\textit{resp.}, $\hat{M}_t$) and reference frame $\hat{X}_{t-1}^{mi}$ (\textit{resp.}, $\hat{X}_{t-1}$) from the frame buffer, we can obtain the predicted frame $\overline{X}_{t}^{mi}$ (\textit{resp.}, $\overline{X}_t$) using the motion compensation module. The residual between input frame $X_t$ and the predicted frame $\overline{X}_t$ is then compressed to $\hat{R}_t^{mi}$ or $\hat{R}_t$ by our adaptive residual compression module. Finally, by adding $\hat{M}_t^{mi}$ (\textit{resp.}, $\hat{M}_t$) with $\hat{R}_t^{mi}$ (\textit{resp.}, $\hat{R}_t$), we can get the reconstructed frame $\hat{X}_{t}^{mi}$ (\textit{resp. $\hat{X}_{t}$}) for $i$-th machine vision task (\textit{resp.}, human vision).

In ``EAC (NVC)'', to simultaneously balance the optimizations for multiple machine vision tasks and human vision, we utilize the adaptive compression method in motion compression networks, namely adaptive motion compression.
During the adaptive motion compression process, the estimated motion $M_t$ is encoded and quantized into the feature $\mathrm{\hat{y}_t^{mot}}$. 
At the same time, we predict binarized masks through our adaptive predictors, where we leverage the hyperprior information produced by the hyperprior network as the spatial information and the feature extracted from the previously reconstructed frame $\hat{X}_{t-1}^{mi}$ as the temporal information. We fuse these two pieces of information using standard convolutional operations followed by a max or Gumbel-softmax~\cite{gumbel} operation to produce binarized masks in predictors.
With the aid of these masks, we adopt the adaptive compression method described in Section~\ref{section:split}, compressing the quantized motion feature $\mathrm{\hat{y}_t^{mot}}$ into different motion features $\mathrm{\hat{y}_t^{mot1}}, ..., \mathrm{\hat{y}_t}$. Then, $\mathrm{\hat{y}_t^{mot1}}$ is reconstructed for the first machine vision task. Similarly, $\mathrm{\hat{y}_t^{moti}}, i\in\{2,3,...\}$ is reconsturcted for the $i$-th vision tasks, respectively. $\mathrm{\hat{y}_t^{mot}}$ is decoded for human vision.
Similarly, our adaptive residual compression adopts an identical network architecture as the adaptive motion compression to adaptively compress the residual information $R_t$ into quantized residual features $\mathrm{\hat{y}_t^{resi}}, i\in\{1,2,...\}$ (\textit{resp.}, $\mathrm{\hat{y}_t^{res}}$) and reconstruct residual $\hat{R}_t^{mi}$ (\textit{resp.}, $\hat{R}_t$) for the $i$-th machine vision task (\textit{resp.}, human vision).
In addition, like ``EAC (NIC)'', our ``EAC (NVC)'' also adopts task-specific adapters to optimize all task-specific networks effectively. But different with ``EAC (NIC)'', the adapters in ``EAC (NVC)'' additionally introduce temporal information from multiple reference frames (\textit{i.e.}, $\hat{X}_{t-1}^{mi}, \hat{X}_{t-2}^{mi}, \hat{X}_{t-3}^{mi}$) to more effectively stimulate the task-specific network, as described in Section~\ref{section:adapter}.

\subsection{Optimization}
\label{section:optimization}
We optimize both ``EAC (NIC)'' and ``EAC (NVC)'' in a two-stage training procedure. 

\subsubsection{\textbf{In the First Stage}} We optimize predictors for adaptive compression module.
For ``EAC (NIC)'', each predictor is trained individually and the predictor $i(i\in \{1,2,...,n\})$ will be optimized as below,
\vspace{-2mm}
\begin{equation}
\label{eq:loss_image}
 \mathcal{L}_{p} = \mathcal{R}(\hat{y}_i) + \lambda \times {\mathcal{D}(f_\phi(\hat{x}_i), G_t)},
\vspace{-2mm}
\end{equation}
$f_\phi$ is the task-special network (\textit{e.g.}, ResNet50~\cite{resnet} for the classification task), and its parameters are frozen during the whole training process. $\hat{x}_i$ is the image reconstructed by the compression modules and used for the $i$-th machine vision task. $G_t$ is the ground-truth for the machine vision task. The function $\mathcal{D}(\cdot)$ provided by the task-special network calculates the loss between $f_\phi(\hat{x}_i)$ and $G_t$. $\hat{y}_i$ is the quantized latent representation used for the $i$-th machine vision task. $\mathcal{R}(\hat{y}_i)$ represents bit-rate costs of $\hat{y}_i$. $\lambda$ is a hyper-parameter, which balance $\mathcal{R}(\hat{y}_i)$ and $\mathcal{D}(f_\phi(\hat{x}_{i}), G_t)$ during the training process.

For ``EAC (NVC)'', our adaptive compression mechanisms (\textit{i.e.}, adaptive motion compression module and adaptive residual compression module) will cause severe accumulative errors for machine-vision-based reconstruction due to only a subset of quantized motion and residual features being encoded and transmitted.  
To mitigate this, we then adopt a cumulative objective function as in~\cite{lin2020mlvc} to optimize adaptive motion and residual compression modules during the training stage, in which we back-propagate the sum of rate-distortion (RD) values produced by compressing $T$ consecutive video frames.
It is also worth mentioning that in this stage, we only optimize the adaptive predictors (\textit{e.g.}, Preditor $i$) in the adaptive motion and residual compression network by resolving the following objective function:
\vspace{-1mm}
\begin{equation}
    % \mathcal{L}_{p}^{*} = \frac{1}{T}\sum_{t}^{T}(\underbrace{\mathcal{R}(\hat{M}_t^{mi})\!+\!\mathcal{R}(\hat{R}_t^{mi})}_{\mathcal{L}_{bit}} + \\ \lambda \underbrace{\mathcal{D}(f_{\phi}(\hat{X}_t^{mi}), G_t)}_{\mathcal{L}_{machine}}),
    % \label{loss_video}
    \mathcal{L}_{p}^{*} = \frac{1}{T}\sum_{t}^{T}(\mathcal{R}(\hat{M}_t^{mi})\!+\!\mathcal{R}(\hat{R}_t^{mi}) + \\ \lambda \times \mathcal{D}(f_{\phi}(\hat{X}_t^{mi}), G_t),
    \label{loss_video}
\vspace{-1mm}
\end{equation}
where $\mathcal{D}(\cdot)$ is the optimization term for the specific downstream tasks, $f_\phi$ denotes the task-specific network (\textit{e.g.}, TSN~\cite{tsn}). $G_t$ is the corresponding ground-truth and $\hat{X}_t^{mi}$ is the reconstructed frame for the $i$-th machine vision task at the $t$-th time-step.
Meanwhile, $\mathcal{R}(\cdot)$ denotes the bit-rate costs respectively for the reconstructed motion $\hat{M}_t^{mi}$ and residual $\hat{R}_t^{mi}$.
Last we use a hyper-parameter $\lambda$ to balance bit-rate costs (\textit{i.e}., $\mathcal{R}(\hat{M}_t^{mi})$ and $\mathcal{R}(\hat{R}_t^{mi})$) and machine vision loss (\textit{i.e.}, $\mathcal{D}(f_{\phi}(\hat{X}_t^{mi}), G_t)$). 

\subsubsection{\textbf{In the Second Stage}} We optimize the $i$-th task-specific adapter to stimulate the $i$-th task-specific network following:
\begin{equation}
\begin{aligned}
\vspace{-1mm}
\label{eq:adapter}
    \mathcal{L}_i = \mathcal{D}(P_i, G_t)
\end{aligned}
\vspace{-1mm}
\end{equation}
%where $\mathcal{D}(\cdot)$ is the task-specific loss function adopted in the downstream task and $G_t$ is the ground-truth for the machine vision task. 
$P_i$ is the prediction for the $i$-th downstream task. For ``EAC (NIC)'', we adopt $P_i$ from Eq.~\ref{eq:adapter_image}. For ``EAC (NVC)'', we utilize $P_i$ from Eq.~\ref{eq:adapter_video}.

\begin{figure*}[htbp]
    \centering
    % \includegraphics[width=\linewidth]{}
    % \subfloat[ ]   % 第一张子图的下标（注意：注释要写在[]中括号内）
  {
      \label{fig:subfig1}\includegraphics[width=0.32\textwidth]{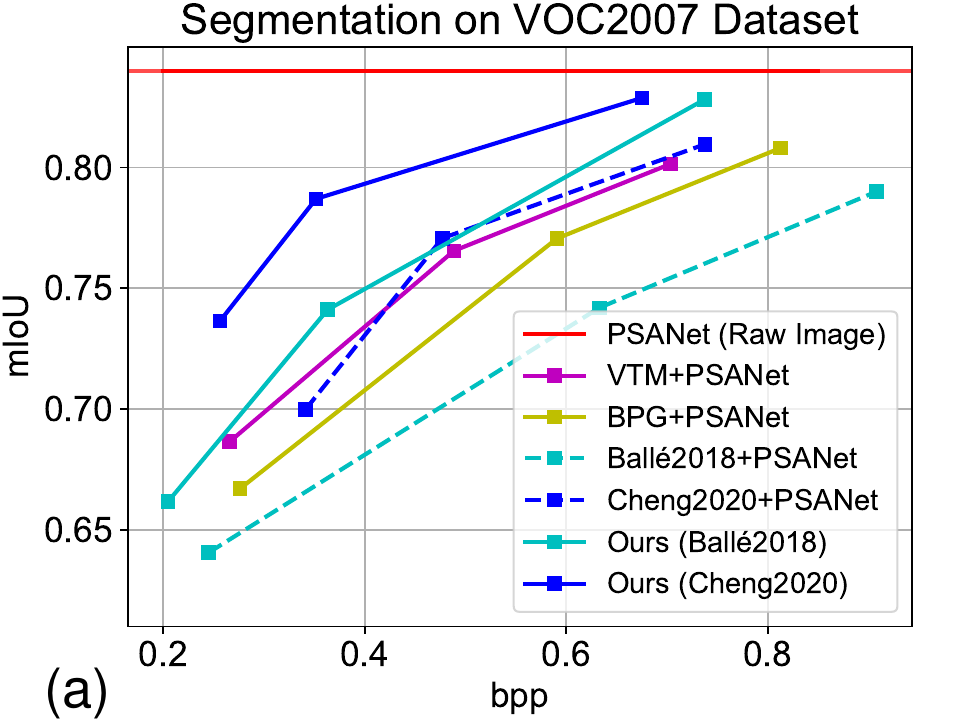}
  }
  % \subfloat[ ]
  {
      \label{fig:subfig2}\includegraphics[width=0.32\textwidth]{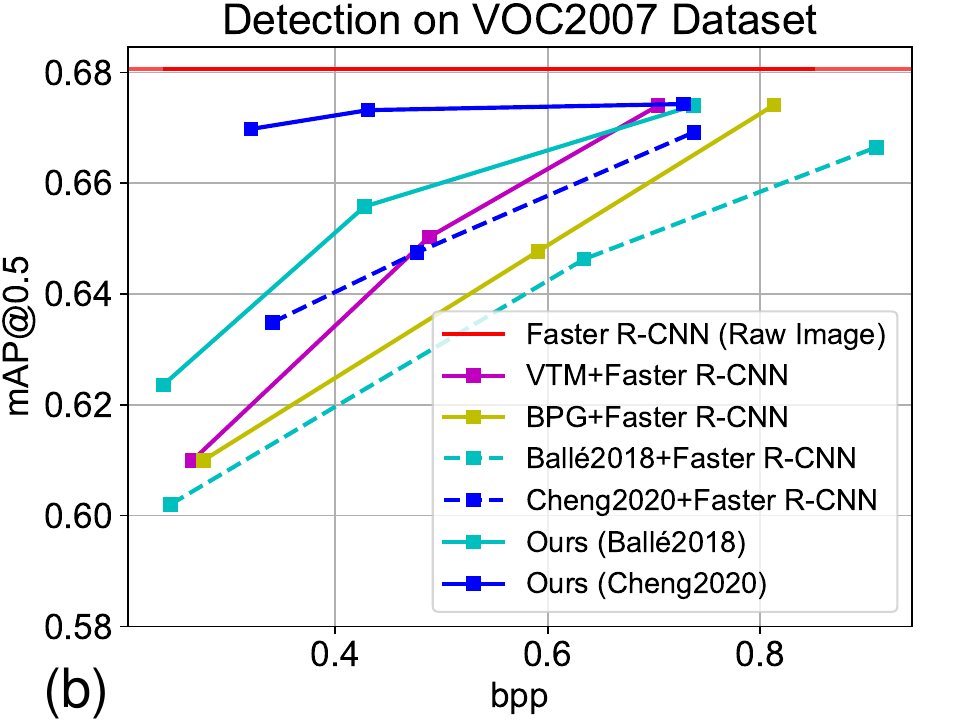}
  }
  % \subfloat[ ]
  {
      \label{fig:subfig3}\includegraphics[width=0.32\textwidth]{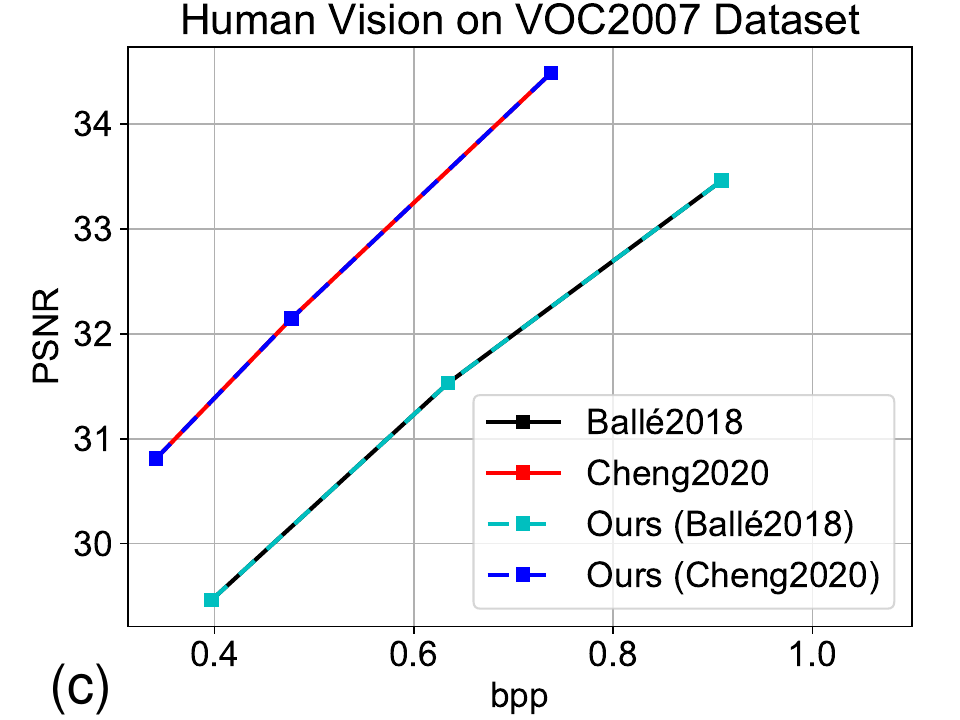}
  }
  \vspace{-7mm}
    \caption{The multi-tasks (\textit{i.e.}, the segmentation task, the detection task, and the human vision) results for our ``EAC (NIC)'' compared to the baseline methods on the VOC2007 dataset. We report mIoU, mAP@0.5, and PSNR results. For all codecs, we use the PSANet as the segmentation network and the Faster R-CNN as the detection network. ``Ours (Ballé2018)''  and ``Ours (Cheng2020)'' denote our ``EAC (NIC)'' framework with Ballé2018 and Cheng2020 as image coding backbone, respectively. ``codec+PSANet'' and ``codec+Faster R-CNN'' denote we directly adopt the codec to compress the images and use such compressed image to perform the segmentation task and the detection task, respectively.}
    \label{fig:image_multi_tasks}
    \vspace{-3mm}
\end{figure*}

\begin{figure*}[htbp]
    \centering
    % \includegraphics[width=\linewidth]{}
  % \subfloat[]
  {
      \label{fig:subfig3}\includegraphics[width=0.32\textwidth]{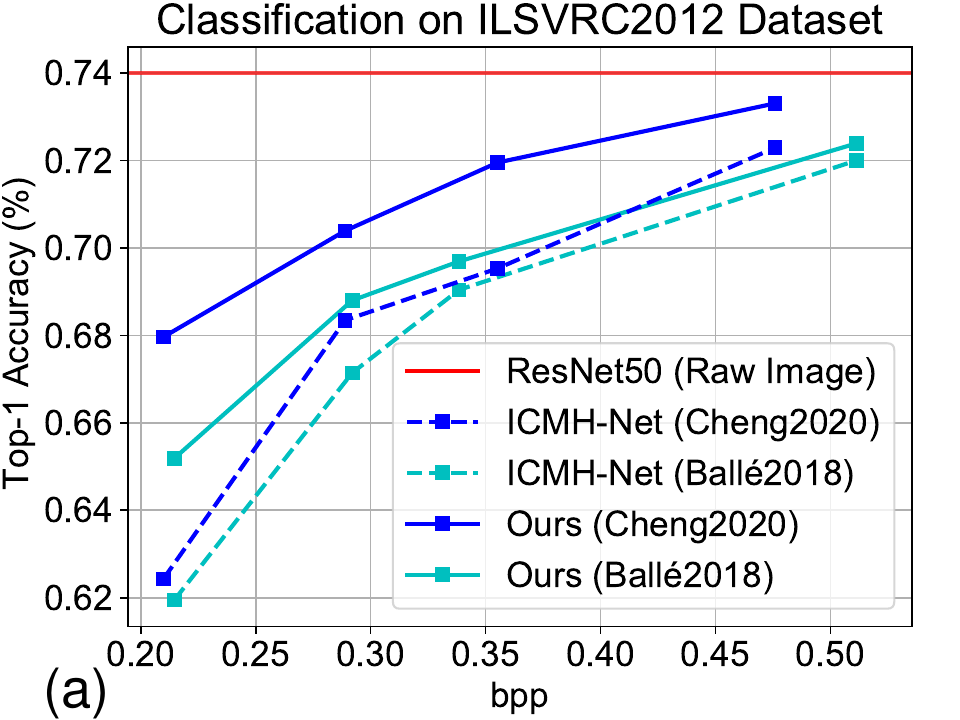}
  }
    % \subfloat[]   % 第一张子图的下标（注意：注释要写在[]中括号内）
  {
      \label{fig:subfig1}\includegraphics[width=0.32\textwidth]{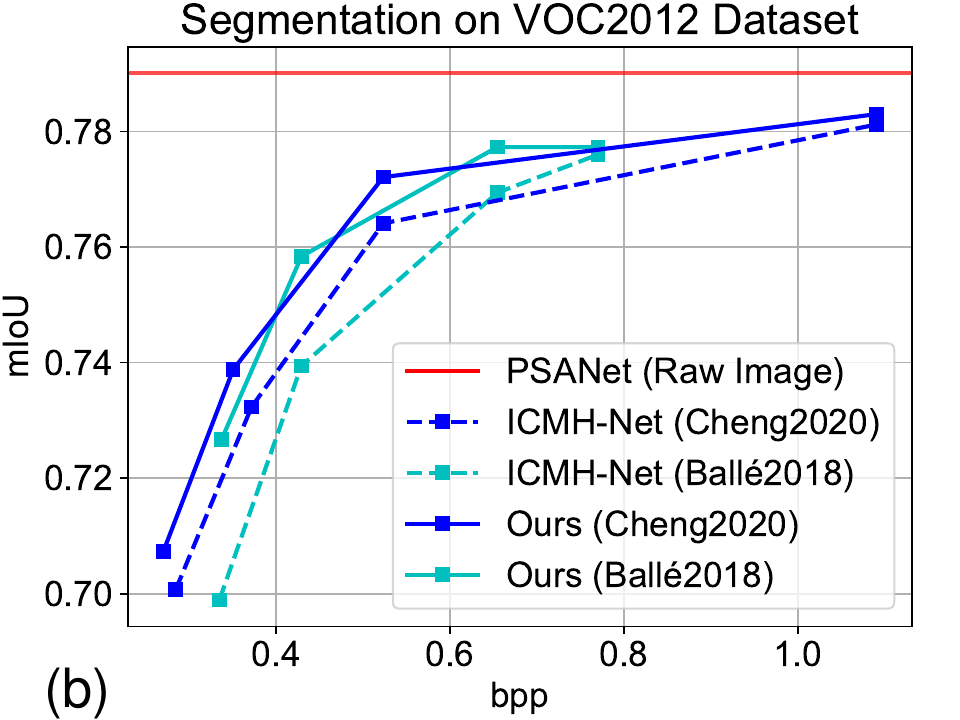}
  }
  % \subfloat[]
  {
      \label{fig:subfig2}\includegraphics[width=0.32\textwidth]{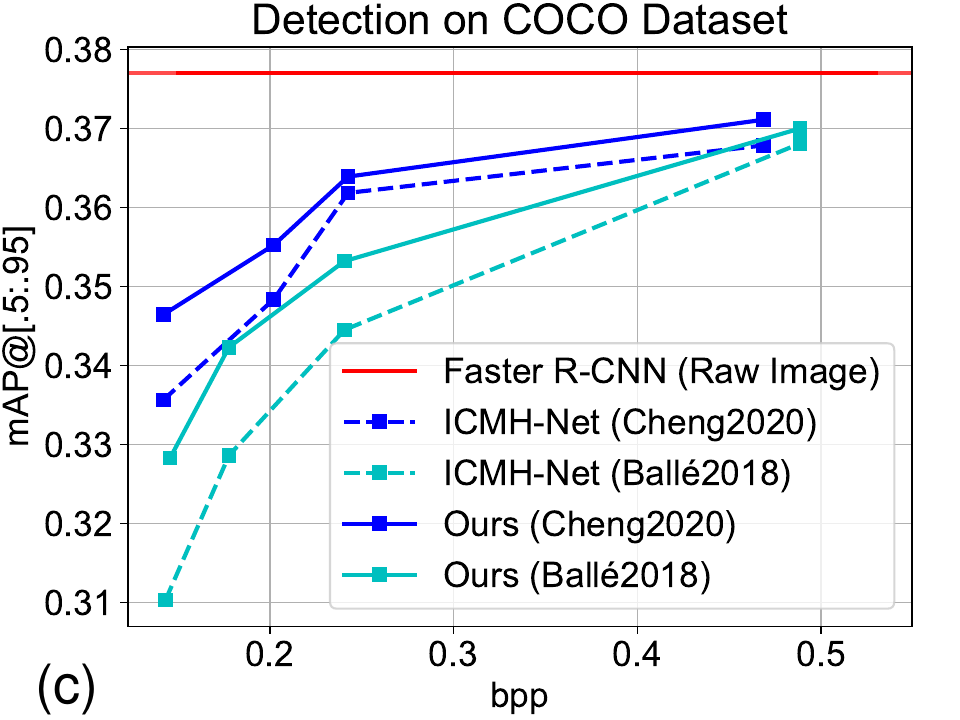}
  }
  \vspace{-7mm}
    \caption{The machine vision tasks (\textit{i.e.}, the classification task, the segmentation task, and the detection task) results for our ``EAC (NIC)'' compared to ICMH-Net on the ILSVRC2012, VOC2012, and COCO dataset. We report accuracy, mIoU, and mAP results. For all codecs, we use the ResNet50 as the classification network, PSANet as the segmentation network, and the Faster R-CNN as the detection network. ``Ours (codec)'' or ``ICMH-Net (codec)'' denotes our ``EAC (NIC)'' framework or ICMH-Net with codec (\textit{i.e.}, Ballé2018, and Cheng2020) as image coding backbone, respectively. }
    \label{fig:image_single_tasks}
    \vspace{-3mm}
\end{figure*}

\section{Experiment}
% \vspace{-1mm}
\subsection{Dataset}
We use ILSVRC 2012~\cite{deng2009imagenet}, PASCAL VOC~\cite{voc}, COCO~\cite{coco} for our ``EAC (NIC)'', and use UCF101~\cite{soomro2012ucf101}, and DAVIS2017~\cite{perazzi2016davis} for our ``EAC (NVC)''. Each dataset adopts the same training/testing splits as the corresponding task-specific networks (\textit{i.e.}, ResNet, PSANet, Faster R-CNN, TSN, and XMem).

% \textbf{ILSVRC 2012.} ILSVRC 2012 dataset~\cite{deng2009imagenet}, also named ImageNet-1k, is a widely used benchmark for the image classification task. ILSVRC 2012 dataset contains over 1.2 million images of 1,000 classes for training. We used the validation set of 50,000 images for testing. 

% \textbf{PASCAL VOC.} PASCAL VOC dataset~\cite{voc} contains VOC 2007 and VOC 2012 datasets, which are designed for both segmentation task and detection task. For the segmentation task, the VOC 2007 dataset has 422 images for training and 210 images for testing, while the VOC 2012 dataset contains 1464 training images and 1449 testing images. Both VOC 2007 and VOC 2012 datasets contain 20 object categories.

% \textbf{COCO.} COCO dataset~\cite{coco} is a large-scale dataset with 80 object categories widely used for the object detection task. The dataset has 118,000 images for training. We used the validation set of 5,000 images for testing. 

% \textbf{UCF101.}
% UCF101~\cite{soomro2012ucf101} is a human actions-focused dataset that we utilize for the task of video action recognition. It encompasses over 13,000 clips distributed into 101 different action categories. We adopt the same training/testing splits following TSN~\cite{tsn}.

% \textbf{DAVIS.}
% DAVIS2017~\cite{perazzi2016davis} is a dataset for video object segmentation. Following XMem~\cite{xmem}, we use 60 videos for training and 30 videos for validation.
\subsection{Experiment Details}
% \vspace{-1mm}
\textbf{Baseline Methods.}
In machine vision tasks, we use compressed images derived from Ballé2018~\cite{ballevariational} or Cheng2020~\cite{cheng2020learned} as input for the task-specific networks in ``EAC (NIC)'' and compressed videos derived from DVC~\cite{dvc} or FVC~\cite{fvc} as input for the task-specific networks in ``EAC (NVC)''. 
For the task-specific networks using compressed images, ResNet~\cite{resnet} with 50 layers is used for the classification task, PSANet~\cite{zhao2018psanet} is adopted for the segmentation task, and Faster R-CNN~\cite{frcnn} is applied to the detection task. For the task-specific networks using compressed videos, we utilize TSN~\cite{tsn} for the object action recognition task and XMem~\cite{xmem} for the object segmentation task. Importantly, we label baseline methods according to the format ``compression method + task-specific network''. For instance, ``FVC+TSN'' denotes adopting the compressed video from FVC as input to TSN. We directly use pre-trained parameters for the neural image/video compression network and the task-specific networks. 
For the conventional codecs, we adopt VTM~\cite{VVC} in the 19.2 version with \textit{lowdelay\_P\_main} configuration setting. Following DeepSVC~\cite{lin2023deepsvc}, we test the x.265 under \textit{veryslow} preset.

\begin{figure*}[!t]%[htbp]
\begin{center}
% \includegraphics[width=0.7\linewidth]{experiment/recognition.pdf}
  % \subfloat[]   % 第一张子图的下标（注意：注释要写在[]中括号内）
  {
      \label{fig:subfig1}\includegraphics[width=0.32\textwidth]{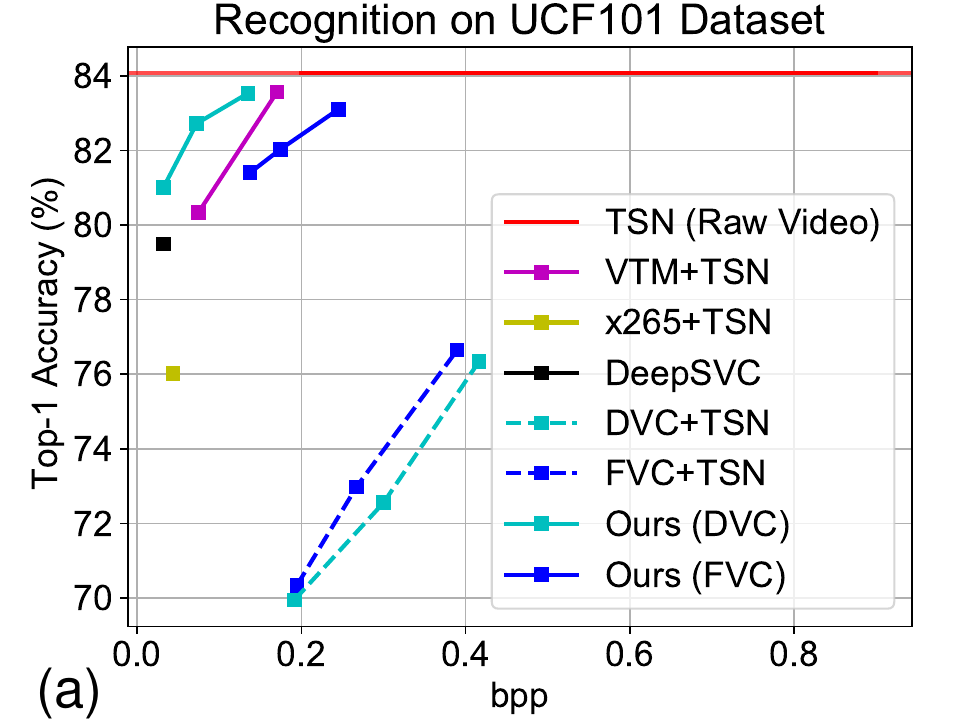}
  }
  % \subfloat[]
  {
      \label{fig:subfig2}\includegraphics[width=0.32\textwidth]{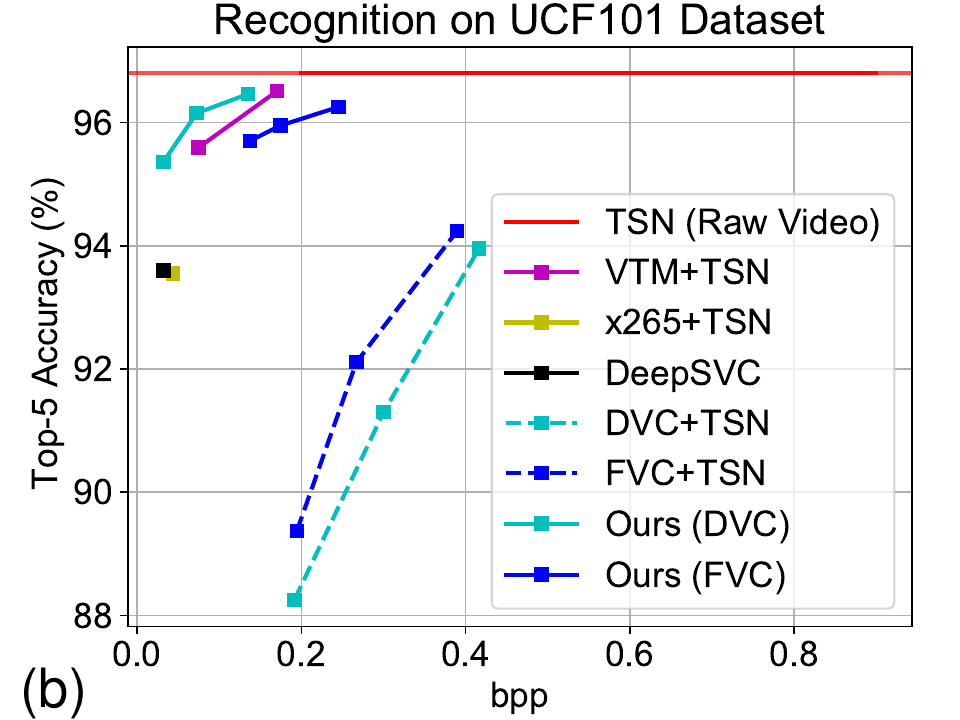}
  }
  {
      \label{fig:subfig3}\includegraphics[width=0.32\textwidth]{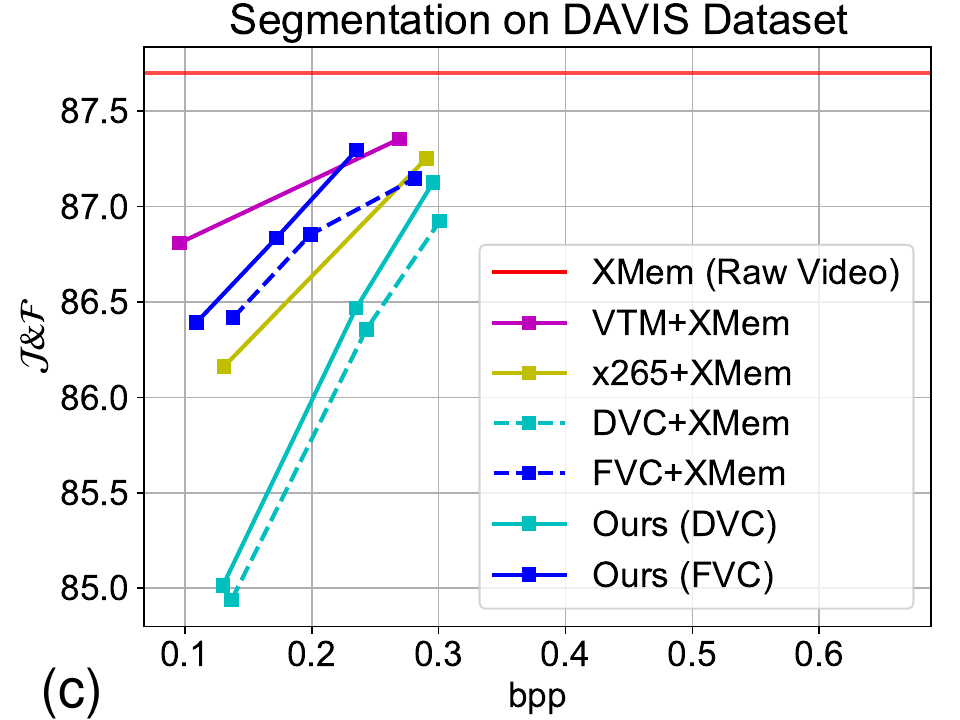}
  }
\end{center}
\vspace{-6mm}
  \caption{Video action recognition results (a), (b) and video object segmentation results (c) for our ``EAC (NVC)'' compared to the baseline methods on the UCF101 dataset. We report both Top-1 accuracy, Top-5 accuracy results for video action recognition, and report the average of Jaccard index and contour accuracy $\mathcal{J\&F}$ results for video object segmentation. For all codecs, we use the TSN as the action recognition network as in DeepSVC, and use the XMem as the video object segmentation network. ``Ours (DVC)'' and ``Ours (FVC)'' denote our ``EAC (NVC)'' framework with DVC and FVC as video coding backbones, respectively. ``codec+TSN" or ``codec+XMem'' denotes we directly adopt the codec to compress the videos and use such compressed videos to perform action recognition task or object segmentation task, respectively.}
  \vspace{-6mm}
\label{fig:tsn}
\end{figure*}

% \begin{figure*}[!t]%[htbp]
% \begin{center}
% % \includegraphics[width=\linewidth]{experiment/segmentation.pdf}
%   % \subfloat[]   % 第一张子图的下标（注意：注释要写在[]中括号内）
%   {
%       \label{fig:subfig1}\includegraphics[width=0.32\textwidth]{experiment/J-Mean.pdf}
%   }
%   % \subfloat[]
%   {
%       \label{fig:subfig2}\includegraphics[width=0.32\textwidth]{experiment/F-Mean.pdf}
%   }
%   % \subfloat[]
%   {
%       \label{fig:subfig3}\includegraphics[width=0.32\textwidth]{experiment/J&F-Mean.pdf}
%   }

% \end{center}
% \vspace{-8mm}
%   \caption{Video object segmentation results for our ``EAC (NVC)'' compared with our baseline methods on the DAVIS dataset. We report Jaccard index $\mathcal{J}$, contour accuracy $\mathcal{F}$, and their average $\mathcal{J\&F}$ results. For all codecs, we use the XMem as the object detection network. ``Ours (DVC)'' and ``Ours (FVC)'' denote our ``EAC (NVC)'' framework with DVC and FVC as video coding backbones, respectively. ``codec+XMen" denotes we directly adopt the codec to compress the videos and use such compressed videos to perform video object segmentation task.}
% \vspace{-5mm}
% \label{fig:xmem}
% \end{figure*}

\textbf{Evaluation Metric.} We use bits per pixel (bpp) to evaluate the bit-rate cost for the compression procedure. We use accuracy, mean intersection-over-union (mIoU), and mean average precision (mAP) to measure the performance for image classification, segmentation, and detection, respectively. Additionally, we evaluate mAP for IoU $\in[0.5: 0.05: 0.95]$ (denoted as mAP@[.5, .95]) which is COCO~\cite{coco}'s standard metric and mAP for IoU $=0.5$ (denoted mAP@0.5) which is PASCAL VOC~\cite{voc}'s metric. We use accuracy (\textit{i.e.}, Top-1 accuracy, and Top-5 accuracy) to evaluate the machine vision performance on the video action recognition task. For the video object segmentation task, we adopt the standard metrics~\cite{perazzi2016benchmark}: the average between Jaccard index and contour ($\mathcal{J\&F}$) to measure the performance. We use the peak signal-to-noise ratio (PSNR) to measure human vision performance.

{\textbf{Implementation Details.}}
We optimize ``EAC (NIC)'' (\textit{resp.}, ``EAC (NVC)'') in a two-stage training strategy. In Stage I, we train our predictors using the loss function in Eq.~\ref{eq:loss_image} (\textit{resp.}, Eq.~\ref{loss_video}) without adopting adapter in ``EAC (NIC)'' (\textit{resp.}, ``EAC (NVC)''). 
In Stage II, we only train the adapter while freezing the parameters in other modules for the ``EAC (NIC)'' (\textit{resp.}, the ``EAC (NVC)'').
Throughout the entire training process, for the sake of ensuring fair comparison, we persistently employ the same pre-trained encoder, decoder, hyperprior network, and task-specific network as utilized in our baseline methods in both ``EAC (NIC)'' and ``EAC (NVC)'', with an assurance that their parameters remain fixed.
Our framework is implemented by PyTorch with CUDA support. All experiments are conducted on the machine with NVIDIA RTX 3090 GPU (24GB memory) for different machine vision tasks. 
We use the Adam optimizer~\cite{kingma2015adam} with the learning rates of 1e-4 (\textit{resp.}, 1e-3) for the first 4 epochs and 1e-5 (\textit{resp.}, 1e-4) for the last 1 epoch in the first (\textit{resp.}, second) stage. We set the batch size as 96 in the ``EAC (NIC)'' and adopted the batch size as 8 in ``EAC (NVC)''. 
% For the image compression branch, we set the batch size as 128 and use the Adam optimizer~\cite{kingma2015adam}
% For the video compression branch, we set the batch size as 8 and use the Adam optimizer~\cite{kingma2015adam} with the learning rates of 1e-4 (\textit{resp.}, 1e-3) for the first 4 epochs and 1e-5 (\textit{resp.}, 1e-4) for the last 1 epoch in the first (\textit{resp.}, second) stage.

\subsection{Experiment Results}
% \vspace{-2mm}
\textbf{Machine Vision Results of ``EAC (NIC)''.}
We provide the RD curves of different compression methods in Fig.~\ref{fig:image_multi_tasks} for multiple tasks (\textit{i.e.}, the segmentation task, the detection task, and the human vision task), it is noted that our method outperforms the baseline method (\textit{i.e.}, ``Ballé2018+PSANet'', and ``Ballé2018+Faster R-CNN'') and other image compression methods (\textit{i.e.}, VTM and BPG) in most bit-rate. For example, our method saves more than 33\% bit-rate than ``Ballé2018+PSANet'' at 0.79 mIoU and saves more than 33\% bit-rate than ``Ballé2018+Faster R-CNN'' at about 0.67 mAP@0.5. The experiment results provide that the effectiveness of our proposed method. Additionally, we provide the RD curves of different compression methods in Fig.~\ref{fig:image_single_tasks} for single machine vision task with only one predictor, which provides that our method ``Ours (Ballé2018)'' (\textit{resp.}, ``Ours (Cheng2020)'') surpasses our conference version ``ICMH-Net (Ballé2018)'' (\textit{resp.}, ``ICMH-Net (Cheng2020)'') on the ILSVRC2012, VOC2012, and COCO datasets. The experiment results in Fig.~\ref{fig:image_single_tasks} also demonstrate the effectiveness of our method in image compression.

\textbf{Machine Vision Results of ``EAC (NVC)''.}
We provide the RD curves of different compression methods in Fig.~\ref{fig:tsn}. It is note that our method ``Ours (DVC)'' (\textit{resp.}, ``Ours (FVC)'') outperforms the baseline method ``DVC+TSN'' and ``DVC+XMem'' (\textit{resp.}, ``FVC+TSN'' and ``FVC+XMem''). For example, ``Ours (DVC)'' improves about 14\% accuracy than ``DVC+TSN'' at 0.2 bpp. Additionally, ``Ours (DVC)'' has better performance than all conventional codecs (\textit{i.e.}, VTM, and x265) in the video action recognition task. The experiment results demonstrate the effectiveness of our method in video compression. 

\textbf{Human Vision Results.}
As we adopt the same pre-trained codecs as our baseline method, and use the full latent feature for human vision. So the human vision performance in our methods (\textit{e.g.}, ``Ours (Cheng2020)'') is similar to our baseline methods (\textit{e.g.}, ``Cheng2020''). It is worse noting that some recent coding for machine methods~\cite{liu2021semantics,torfason2018towards,song2021variable,wang2021end,mei2021learn,9414465,liu2023icme} sacrifice human vision performance while improving the machine vision results. Therefore, the
results indicate the advantage of our method can improve the performance
for the machine vision tasks while maintaining the human vision performance.

% we stack two predictors for multi-tasks (\textit{e.g.}, the segmentation task, the detection task, and the human vision task), as shown in Figure~\ref{fig:multi_exp}. For multi-tasks, the latent representation used for the first machine vision task (\textit{e.g.}, the segmentation task) will be reused for the second machine vision task (\textit{e.g.}, the detection task). And human vision will reuse the latent representations which are used for the first and second machine vision tasks. In Figure~\ref{fig:multi_exp}, our method saves about 25\% bpp at 0.70 mIoU and improves about 5\% mIoU at 0.63 bpp for the first machine vision task. For the second machine vision task, our method achieves about 4\% improvement at the 0.4 bpp and saves about 16\% bpp at the highest mAP@0.5, as shown in Figure~\ref{fig:multi_exp} (b). For human vision, our method also has the same performance as our baseline method, as shown in Figure~\ref{fig:multi_exp} (c). The experiment results demonstrate that our proposed method can improve the performance for multiple machine vision tasks (\textit{i.e.}, the segmentation task and the detection task) without sacrificing the human vision performance. The detail of our ICMH-Net for multi-tasks could be found in the supplementary.

\subsection{Model Analysis}

\begin{figure}[!t]
\vspace{3mm}
\centering
    {
    \includegraphics[width=0.48\linewidth]{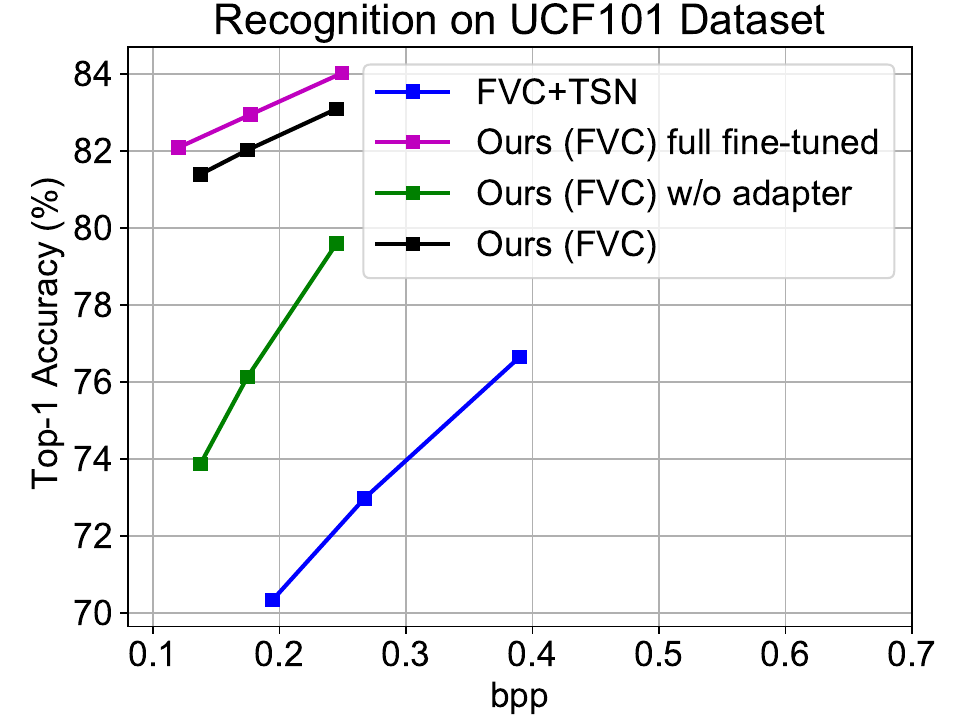}
    }
    {
    \includegraphics[width=0.48\linewidth]{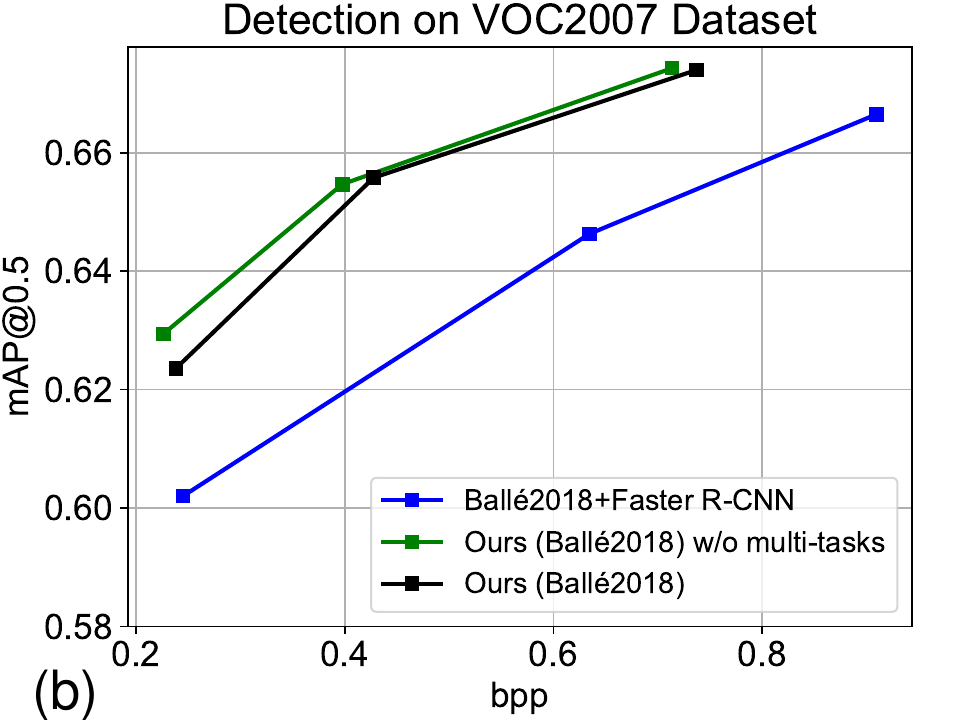}
    }
  \vspace{-7mm}
  \caption{(a) Ablation study of our adaptive motion and residual compression network and task-specific adapter on the UCF101 dataset, where we use FVC as our video coding backbone. (1) \textbf{``FVC+TSN''}: our baseline methods, where we simply compressed the videos and used them for subsequent machine vision tasks (\textit{i.e.}, TSN) with FVC. (2) \textbf{``Ours (FVC) full fine-tuned''}: our alternative method using the full fine-tuned task-specific network. 
  (3) \textbf{``Ours (FVC) w/o adapter''}: our alternative method without adopting the task-specific adapter for the task-specific network. (4) \textbf{``Ours (FVC)''}: our proposed complete method using both adapter and adaptive compression module. (b) Ablation study for multiple machine vision tasks (\textit{i.e.} segmentation, detection) on the VOC2007 dataset, where we use ``Ballé2018'' as our video coding backbone. (1) \textbf{``Ballé2018+Faster R-CNN''}: our baseline method, where we compressed the images and used them for subsequent machine vision tasks (\textit{i.e.}, Faster R-CNN). (2) \textbf{``Ours w/o multi-tasks''}: our alternative method only optimized for a signal vision task (\textit{i.e.}, detection). (3) \textbf{``Ours (Ballé2018)''}: our proposed complete method optimized for multiple machine vision tasks. }
\label{fig:ab}
\vspace{-5mm}
\end{figure}

\begin{table*}[t]
% \footnotesize
\centering
\caption{The complexity (in \#parameters) of our method using task-specific adapter to delta-tune the task-specific network (\textit{i.e.}, TSN, XMem, and ResNet50 in ``EAC (NIC)''. PSANet, and Faster R-CNN in ``EAC (NVC)'') and directly fine-tuning the task-specific network.}
\vspace{-3mm}
\begin{tabular}{c|cccc|ccc}
\hline
Module  & ResNet50 & PSANet & Faster R-CNN & Adapter in ``EAC (NIC)''  & TSN  & XMem  &Adapter in ``EAC (NVC)'' \\
\#Parameters (M)      & 25.56 & 69.80  & 48.30  & \textbf{0.17} & 10.37  & 62.19 &\textbf{0.20}\\ 
\hline
% Module in VCB     & TSN  & XMem  &  &Adapter  \\ 
% \#Param (M)      & 10.37  & 62.19  & &\textbf{0.20}     \\   
% \hline
\end{tabular}
\vspace{-5mm}
\label{Table:parameter}
\end{table*}

\begin{table*}[t]
% \footnotesize
\centering
\caption{The Running Speed of codecs VTM, BPG, Ballé2018, Cheng2020, and our methods ``Ours (Ballé2018)'' and ``Ours (Cheng2020)'' on the images with a resolution of 512x512 pixels from the VOC2007 dataset.}
\vspace{-3mm}
\begin{tabular}{c|cccccc}
\hline
    & VTM & BPG & Ballé2018  &  Ours (Ballé2018)   & Cheng2020 & Ours (Cheng2020) \\
 Running Speed      & 0.021 fps & 1.61 fps  & 155.97 fps & 139.31 fps  & 34.11 fps & 33.30 fps \\ 
\hline

\end{tabular}
\vspace{-7mm}
\label{Table:speed}
\end{table*}

\begin{table}[t]
% \footnotesize
\centering
\caption{Ablation study for effectiveness of using temporal information (\textit{i.e}., reference frame) in our adaptive compression module. ``EAC (NVC)'' uses FVC as the video coding backbone and is tested on the UCF101 dataset. (1) \textbf{``Ours (FVC)*  w/o Reference''}: our alternative method without using temporal information in the adaptive compression module.  (2) \textbf{``Ours (FVC)*''}: our method with complete adaptive compression module. ``*'' means our framework without adopting the task-specific adapter.}
\vspace{-3mm}
\begin{tabular}{ccc}
\hline
Method       & bpp        & Top-1 Accuracy(\%)     \\ \hline
Ours (FVC)* w/o Temporal Information       & 0.14       & 73.22         \\
Ours (FVC)*          & 0.14      & \textbf{73.87}        \\ \hline
\end{tabular}
\vspace{-5mm}
\label{Table:wo_ref}
\end{table}

\textbf{Effectiveness of Different Components.}
The ablation study of our proposed method is shown in Fig.~\ref{fig:ab}. 
As shown in Fig.~\ref{fig:ab} (a), to demonstrate the effectiveness of the parameter-efficient delta-tuning strategy, we remove the adapter in our method and the results of ``Ours (FVC) w/o adapter'' drop about 8\% accuracy at 0.14 bpp when compared with our complete method (\textit{i.e.}, ``Ours (FVC)''). To demonstrate the effectiveness of our adaptive compression method, we further remove the adaptive compression method in ``Ours (FVC) w/o adapter'', namely ``FVC+TSN'', and the results of ``FVC+TSN'' drop over 7\% accuracy at 0.24 bpp when compared with ``Ours (FVC) w/o adapter''. These experimental results demonstrate the effectiveness of our adaptive compression method and parameter-efficient delta-tuning strategy for improving machine vision performance. Additionally, the experiment results in Table~\ref{Table:wo_ref} demonstrate the effectiveness when using the temporal information (\textit{i.e.} reference frame) for our adaptive predictor in ``EAC (NVC)''.
% In Fig.~\ref{fig:ab} (a), xxx. In Fig.~\ref{fig:ab} (b), we take ``FVC+TSN'' as our baselines. When compared with ``FVC+TSN'', our method without task-specific adapter achieves 6\% accuracy improvement at 0.2 bpp, which indicates the effectiveness of our adaptive motion and residual compression network. When using the task-specific adapter, our method will further improve 5\% accuracy at 0.2 bpp than our method without the adapter, which provides that our task-specific adapter can improve the machine vision performance.  Additionally, the experiment in Table~\ref{Table:wo_ref} demonstrates the effectiveness when using the reference frame in our adaptive predictor.

To measure the impact of optimizing our adaptive compression method for multiple machine vision tasks, we have constructed an ablation study for adaptive compression, as shown in Fig.~\ref{fig:ab} (b). When our adaptive compression method is optimized for both segmentation and detection tasks simultaneously, its performance on the detection task (see the black curve) drops about 0.0001 mAP at 0.71 bpp when compared with the method optimized only for the detection task (see the green curve). 
% This is because our method reuses latent features, which are utilized by the segmentation task, when used for the detection task, resulting in a bit of extra bit-rate consumption. 
This is because our method reuses latent features from the segmentation task for the detection task, which leads to a slight increase in bit-rate consumption.
Hence, by using the adaptive compression method, our approach can be applied to a variety of machine vision tasks simultaneously (see Fig.~\ref{fig:image_multi_tasks}), with only a slight sacrifice in machine vision performance.

\textbf{Complexity Analysis.}
As shown in Table~\ref{Table:parameter}, our methodology using the task-specific adapter to delta-tune the task-specific network (\textit{i.e.}, TSN, XMem, ResNet50, PSANet, and Faster R-CNN) saves 98.1\%, 99.7\%, 99.3\%, 99.8\%, and 99.6\% parameters compared to directly fine-tune TSN, XMem, ResNet50, PSANet, and Faster R-CNN, respectively. These parameter numbers confirm that our task-specific adapter acts as a more lightweight network compared to the task-specific network.

\textbf{Running Speed.}
We use the images with a resolution of $512\times512$ from the VOC2007~\cite{voc} dataset as an example to evaluate the running speeds of codecs. As shown in Table~\ref{Table:speed}, the running speeds of the conventional codecs VTM~\cite{VVC} and BPG~\cite{BPG} are 0.021fps and 1.61fps, which are much slower than our methods ``Ours (Ballé2018)'' (139.31fps) and ``Ours (Cheng2020)'' (33.30fps). In addition, despite ``Ours (Cheng2020)'' incorporating a predictor compared to the baseline method Cheng2020~\cite{cheng2020learned}, the running speed is only 2\% higher than Cheng2020. Hence, our methods can achieve better machine vision performance (see Fig.~\ref{fig:image_multi_tasks}) with only a slight increase in computation cost.

\begin{figure}[!t]%[htbp]
\begin{center}
\includegraphics[width=0.7\linewidth]{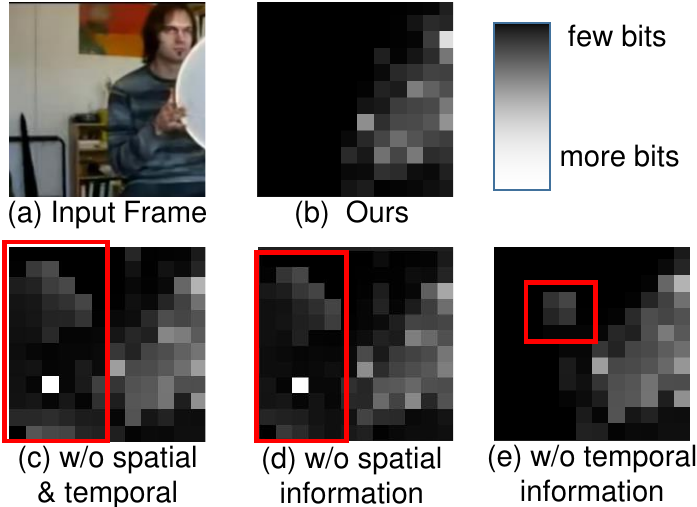}
\end{center}
\vspace{-5mm}
  \caption{Visualization of (a) the input frame, (b) the selected feature map when using both spatial and temporal information, (c) the selected feature map when without using both spatial information and temporal information, (d) the selected feature map without using spatial information, (e) the selected feature map without using temporal information.}
\vspace{-6mm}
\label{fig:visualization}
\end{figure}

\textbf{Quantized Feature Visualization.}
When ``EAC (NVC)'' is used for the video object detection task on the UCF101 dataset, we provide the visualization of our quantized motion features to demonstrate the effectiveness of our adaptive compression method.
The input frame, as shown in Fig.~\ref{fig:visualization} (a), is compressed to the unselected motion feature, 
which costs much bit-rate in the background area, as shown in the red bounding box of Fig.~\ref{fig:visualization} (c), which might be much less important for the object recognition task.
If we only use temporal information (\textit{i.e.}, the reference frames) or spatial information (\textit{i.e.}, the hyperprior information), we can generate a better selection with lower redundancy in this area, as shown in Fig.~\ref{fig:visualization} (c) and (d) respectively.
Moreover, if we adopt both temporal and spatial information in the adaptive motion compression network, we can further improve our feature selection performance by removing the bit-rate cost in that less-important areas (\textit{e.g.}, background) for the object recognition task, as shown in Fig.~\ref{fig:visualization} (b). 
% The visualization results demonstrate that the efficient

\section{Conclusion}
In this work, we propose an efficient adaptive compression (EAC) method for human perception and multiple machine vision tasks. EAC involves an adaptive compression module specifically designed for multiple vision tasks by adaptively selecting subsets from the quantized latent features, and lightweight task-specific adapters to effectively optimize the task-specific networks. Our EAC seamlessly integrates into various NIC methods and NVC methods, extending their capabilities to improve machine perception.
A thorough set of experiments and ablation studies were performed to demonstrate the coding capabilities for multiple machine vision tasks (\textit{e.g.}, segmentation, and detection), resulting in superior performance compared to recent coding for machine methods (\textit{i.e.}, ICMH-Net, DeepSVC) and the traditional codec VTM.
Overall, our work establishes a new baseline for human-machine-vision guided image and video compression, which will facilitate subsequent research in this area.

\bibliographystyle{IEEEtran}
\bibliography{reference}

% Generated by IEEEtran.bst, version: 1.14 (2015/08/26)
\begin{thebibliography}{10}
\providecommand{\url}[1]{#1}
\csname url@samestyle\endcsname
\providecommand{\newblock}{\relax}
\providecommand{\bibinfo}[2]{#2}
\providecommand{\BIBentrySTDinterwordspacing}{\spaceskip=0pt\relax}
\providecommand{\BIBentryALTinterwordstretchfactor}{4}
\providecommand{\BIBentryALTinterwordspacing}{\spaceskip=\fontdimen2\font plus
\BIBentryALTinterwordstretchfactor\fontdimen3\font minus \fontdimen4\font\relax}
\providecommand{\BIBforeignlanguage}[2]{{%
\expandafter\ifx\csname l@#1\endcsname\relax
\typeout{** WARNING: IEEEtran.bst: No hyphenation pattern has been}%
\typeout{** loaded for the language `#1'. Using the pattern for}%
\typeout{** the default language instead.}%
\else
\language=\csname l@#1\endcsname
\fi
#2}}
\providecommand{\BIBdecl}{\relax}
\BIBdecl

\bibitem{liu2021semantics}
K.~Liu, D.~Liu, L.~Li, N.~Yan, and H.~Li, ``Semantics-to-signal scalable image compression with learned revertible representations,'' \emph{{Int. J. Comput. Vision}}, vol. 129, no.~9, pp. 2605--2621, 2021.

\bibitem{torfason2018towards}
R.~Torfason, F.~Mentzer, E.~Agustsson, M.~Tschannen, R.~Timofte, and L.~Van~Gool, ``Towards image understanding from deep compression without decoding,'' \emph{{Proc. Int. Conf. Learn. Representations}}, 2018.

\bibitem{song2021variable}
M.~Song, J.~Choi, and B.~Han, ``Variable-rate deep image compression through spatially-adaptive feature transform,'' in \emph{{Proc. IEEE Int. Conf. Comp. Vis.}}, 2021, pp. 2380--2389.

\bibitem{wang2021end}
S.~Wang, Z.~Wang, S.~Wang, and Y.~Ye, ``End-to-end compression towards machine vision: Network architecture design and optimization,'' \emph{IEEE Open J. Circuits Syst.}, vol.~2, pp. 675--685, 2021.

\bibitem{mei2021learn}
Y.~Mei, F.~Li, L.~Li, and Z.~Li, ``Learn a compression for objection detection-vae with a bridge,'' in \emph{IEEE Int. Conf. Vis. Commun. Image Process.}\hskip 1em plus 0.5em minus 0.4em\relax IEEE, 2021, pp. 1--5.

\bibitem{9414465}
N.~Le, H.~Zhang, F.~Cricri, R.~Ghaznavi-Youvalari, and E.~Rahtu, ``Image coding for machines: an end-to-end learned approach,'' in \emph{{IEEE} Int. Conf. Acoust. Speech Signal Process.}, 2021, pp. 1590--1594.

\bibitem{chen2023transtic}
Y.-H. Chen, Y.-C. Weng, C.-H. Kao, C.~Chien, W.-C. Chiu, and W.-H. Peng, ``Transtic: Transferring transformer-based image compression from human perception to machine perception,'' in \emph{{Proc. IEEE Int. Conf. Comp. Vis.}}, 2023, pp. 23\,297--23\,307.

\bibitem{choi2022scalable}
H.~Choi and I.~V. Baji{\'c}, ``Scalable image coding for humans and machines,'' \emph{{{IEEE} Trans. Image Process.}}, vol.~31, pp. 2739--2754, 2022.

\bibitem{bai2022towards}
Y.~Bai, X.~Yang, X.~Liu, J.~Jiang, Y.~Wang, X.~Ji, and W.~Gao, ``Towards end-to-end image compression and analysis with transformers,'' in \emph{{Proc. Conf. AAAI}}, vol.~36, no.~1, 2022, pp. 104--112.

\bibitem{9385898}
S.~Yang, Y.~Hu, W.~Yang, L.-Y. Duan, and J.~Liu, ``Towards coding for human and machine vision: Scalable face image coding,'' \emph{{{IEEE} Trans. Multimedia}}, pp. 1--1, 2021.

\bibitem{wang2021towards}
S.~Wang, S.~Wang, W.~Yang, X.~Zhang, S.~Wang, S.~Ma, and W.~Gao, ``Towards analysis-friendly face representation with scalable feature and texture compression,'' \emph{{{IEEE} Trans. Multimedia}}, 2021.

\bibitem{lin2023deepsvc}
H.~Lin, B.~Chen, Z.~Zhang, J.~Lin, X.~Wang, and T.~Zhao, ``Deepsvc: Deep scalable video coding for both machine and human vision,'' in \emph{{Proc. {ACM} Int. Conf. Multimedia}}, 2023, pp. 9205--9214.

\bibitem{tian2023non}
Y.~Tian, G.~Lu, G.~Zhai, and Z.~Gao, ``Non-semantics suppressed mask learning for unsupervised video semantic compression,'' in \emph{{Proc. IEEE Int. Conf. Comp. Vis.}}, 2023, pp. 13\,610--13\,622.

\bibitem{ding2023parameter}
N.~Ding, Y.~Qin, G.~Yang, F.~Wei, Z.~Yang, Y.~Su, S.~Hu, Y.~Chen, C.-M. Chan, W.~Chen \emph{et~al.}, ``Parameter-efficient fine-tuning of large-scale pre-trained language models,'' \emph{Natu. Mach. Inte.}, vol.~5, no.~3, pp. 220--235, 2023.

\bibitem{ballevariational}
J.~Ball{\'e}, D.~Minnen, S.~Singh, S.~J. Hwang, and N.~Johnston, ``Variational image compression with a scale hyperprior,'' in \emph{{Proc. Int. Conf. Learn. Representations}}, 2018.

\bibitem{cheng2020learned}
Z.~Cheng, H.~Sun, M.~Takeuchi, and J.~Katto, ``Learned image compression with discretized gaussian mixture likelihoods and attention modules,'' in \emph{{Proc. IEEE Conf. Comp. Vis. Patt. Recogn.}}, 2020, pp. 7939--7948.

\bibitem{dvc}
G.~Lu, W.~Ouyang, D.~Xu, X.~Zhang, C.~Cai, and Z.~Gao, ``Dvc: An end-to-end deep video compression framework,'' in \emph{{Proc. IEEE Conf. Comp. Vis. Patt. Recogn.}}, 2019, pp. 11\,006--11\,015.

\bibitem{fvc}
Z.~Hu, G.~Lu, and D.~Xu, ``Fvc: A new framework towards deep video compression in feature space,'' in \emph{{Proc. IEEE Conf. Comp. Vis. Patt. Recogn.}}, 2021, pp. 1502--1511.

\bibitem{deng2009imagenet}
J.~Deng, W.~Dong, R.~Socher, L.-J. Li, K.~Li, and L.~Fei-Fei, ``Imagenet: A large-scale hierarchical image database,'' in \emph{{Proc. IEEE Conf. Comp. Vis. Patt. Recogn.}}, 2009, pp. 248--255.

\bibitem{voc}
M.~Everingham, L.~Van~Gool, C.~K. Williams, J.~Winn, and A.~Zisserman, ``The pascal visual object classes (voc) challenge,'' \emph{{Int. J. Comput. Vision}}, vol.~88, pp. 303--338, 2010.

\bibitem{coco}
T.-Y. Lin, M.~Maire, S.~Belongie, J.~Hays, P.~Perona, D.~Ramanan, P.~Doll{\'a}r, and C.~L. Zitnick, ``Microsoft coco: Common objects in context,'' in \emph{{Proc. Eur. Conf. Comp. Vis.}}\hskip 1em plus 0.5em minus 0.4em\relax Springer, 2014, pp. 740--755.

\bibitem{soomro2012ucf101}
K.~Soomro, A.~R. Zamir, and M.~Shah, ``Ucf101: A dataset of 101 human actions classes from videos in the wild,'' \emph{arXiv preprint arXiv:1212.0402}, 2012.

\bibitem{perazzi2016davis}
F.~Perazzi, J.~Pont-Tuset, B.~McWilliams, L.~Van~Gool, M.~Gross, and A.~Sorkine-Hornung, ``A benchmark dataset and evaluation methodology for video object segmentation,'' in \emph{{Proc. IEEE Conf. Comp. Vis. Patt. Recogn.}}, 2016, pp. 724--732.

\bibitem{VVC}
``Vvc test model (vtm),'' 2022, \url{https://jvet.hhi.fraunhofer.de/}, accessed:2024.

\bibitem{liu2023icmh}
L.~Liu, Z.~Hu, Z.~Chen, and D.~Xu, ``Icmh-net: Neural image compression towards both machine vision and human vision,'' in \emph{{Proc. {ACM} Int. Conf. Multimedia}}, 2023, pp. 8047--8056.

\bibitem{balle2017end}
J.~Ball{\'e}, V.~Laparra, and E.~P. Simoncelli, ``End-to-end optimized image compression,'' in \emph{{Proc. Int. Conf. Learn. Representations}}, 2017.

\bibitem{minnen2018joint}
D.~Minnen, J.~Ball{\'e}, and G.~D. Toderici, ``Joint autoregressive and hierarchical priors for learned image compression,'' \emph{{Advances in Neural Inf. Process. Syst.}}, vol.~31, 2018.

\bibitem{1}
D.~Wang, W.~Yang, Y.~Hu, and J.~Liu, ``Neural data-dependent transform for learned image compression,'' in \emph{{Proc. IEEE Conf. Comp. Vis. Patt. Recogn.}}, 2022, pp. 17\,379--17\,388.

\bibitem{2}
R.~Zou, C.~Song, and Z.~Zhang, ``The devil is in the details: Window-based attention for image compression,'' in \emph{{Proc. IEEE Conf. Comp. Vis. Patt. Recogn.}}, 2022, pp. 17\,492--17\,501.

\bibitem{3}
X.~Zhu, J.~Song, L.~Gao, F.~Zheng, and H.~T. Shen, ``Unified multivariate gaussian mixture for efficient neural image compression,'' in \emph{{Proc. IEEE Conf. Comp. Vis. Patt. Recogn.}}, 2022, pp. 17\,612--17\,621.

\bibitem{7}
M.~Song, J.~Choi, and B.~Han, ``Variable-rate deep image compression through spatially-adaptive feature transform,'' in \emph{{Proc. IEEE Int. Conf. Comp. Vis.}}, 2021, pp. 2380--2389.

\bibitem{8}
F.~Yang, L.~Herranz, Y.~Cheng, and M.~G. Mozerov, ``Slimmable compressive autoencoders for practical neural image compression,'' in \emph{{Proc. IEEE Conf. Comp. Vis. Patt. Recogn.}}, 2021, pp. 4998--5007.

\bibitem{9}
Y.~Qian, Z.~Tan, X.~Sun, M.~Lin, D.~Li, Z.~Sun, L.~Hao, and R.~Jin, ``Learning accurate entropy model with global reference for image compression,'' in \emph{{Proc. Int. Conf. Learn. Representations}}, 2021.

\bibitem{10}
Y.~Wu, X.~Li, Z.~Zhang, X.~Jin, and Z.~Chen, ``Learned block-based hybrid image compression,'' \emph{{{IEEE} Trans. Circuits Syst. Video Technol.}}, vol.~32, no.~6, pp. 3978--3990, 2021.

\bibitem{11}
Z.~Cui, J.~Wang, S.~Gao, T.~Guo, Y.~Feng, and B.~Bai, ``Asymmetric gained deep image compression with continuous rate adaptation,'' in \emph{{Proc. IEEE Conf. Comp. Vis. Patt. Recogn.}}, 2021, pp. 10\,532--10\,541.

\bibitem{12}
G.~Pan, G.~Lu, Z.~Hu, and D.~Xu, ``Content adaptive latents and decoder for neural image compression,'' in \emph{{Proc. Eur. Conf. Comp. Vis.}}\hskip 1em plus 0.5em minus 0.4em\relax Springer, 2022, pp. 556--573.

\bibitem{he2022elic}
D.~He, Z.~Yang, W.~Peng, R.~Ma, H.~Qin, and Y.~Wang, ``Elic: Efficient learned image compression with unevenly grouped space-channel contextual adaptive coding,'' in \emph{{Proc. IEEE Conf. Comp. Vis. Patt. Recogn.}}, 2022, pp. 5718--5727.

\bibitem{han2024cra5}
T.~Han, Z.~Chen, S.~Guo, W.~Xu, and L.~Bai, ``Cra5: Extreme compression of era5 for portable global climate and weather research via an efficient variational transformer,'' \emph{arXiv preprint arXiv:2405.03376}, 2024.

\bibitem{liu2023icme}
L.~Liu, Z.~Hu, and J.~Zhang, ``Pchm-net: A new point cloud compression framework for both human vision and machine vision,'' \emph{Proc. IEEE Int. Conf. Multimedia Expo}, 2023.

\bibitem{lin2020mlvc}
J.~Lin, D.~Liu, H.~Li, and F.~Wu, ``M-lvc: Multiple frames prediction for learned video compression,'' in \emph{{Proc. IEEE Conf. Comp. Vis. Patt. Recogn.}}, 2020, pp. 3546--3554.

\bibitem{agustsson2020scale}
E.~Agustsson, D.~Minnen, N.~Johnston, J.~Balle, S.~J. Hwang, and G.~Toderici, ``Scale-space flow for end-to-end optimized video compression,'' in \emph{{Proc. IEEE Conf. Comp. Vis. Patt. Recogn.}}, 2020, pp. 8503--8512.

\bibitem{djelouah2019neural}
A.~Djelouah, J.~Campos, S.~Schaub-Meyer, and C.~Schroers, ``Neural inter-frame compression for video coding,'' in \emph{{Proc. IEEE Int. Conf. Comp. Vis.}}, 2019, pp. 6421--6429.

\bibitem{habibian2019video}
A.~Habibian, T.~v. Rozendaal, J.~M. Tomczak, and T.~S. Cohen, ``Video compression with rate-distortion autoencoders,'' in \emph{{Proc. IEEE Int. Conf. Comp. Vis.}}, 2019, pp. 7033--7042.

\bibitem{hu2020improving}
Z.~Hu, Z.~Chen, D.~Xu, G.~Lu, W.~Ouyang, and S.~Gu, ``Improving deep video compression by resolution-adaptive flow coding,'' in \emph{{Proc. Eur. Conf. Comp. Vis.}}\hskip 1em plus 0.5em minus 0.4em\relax Springer, 2020, pp. 193--209.

\bibitem{chen2022exploiting}
Z.~Chen, S.~Gu, G.~Lu, and D.~Xu, ``Exploiting intra-slice and inter-slice redundancy for learning-based lossless volumetric image compression,'' \emph{{{IEEE} Trans. Image Process.}}, vol.~31, pp. 1697--1707, 2022.

\bibitem{hu2022coarse}
Z.~Hu, G.~Lu, J.~Guo, S.~Liu, W.~Jiang, and D.~Xu, ``Coarse-to-fine deep video coding with hyperprior-guided mode prediction,'' in \emph{{Proc. IEEE Conf. Comp. Vis. Patt. Recogn.}}, 2022, pp. 5921--5930.

\bibitem{lu2020content}
G.~Lu, C.~Cai, X.~Zhang, L.~Chen, W.~Ouyang, D.~Xu, and Z.~Gao, ``Content adaptive and error propagation aware deep video compression,'' in \emph{{Proc. Eur. Conf. Comp. Vis.}}\hskip 1em plus 0.5em minus 0.4em\relax Springer, 2020, pp. 456--472.

\bibitem{lu2020end}
G.~Lu, X.~Zhang, W.~Ouyang, L.~Chen, Z.~Gao, and D.~Xu, ``An end-to-end learning framework for video compression,'' \emph{{{IEEE} Trans. Pattern Anal. Mach. Intell.}}, vol.~43, no.~10, pp. 3292--3308, 2020.

\bibitem{yang2020learning}
R.~Yang, F.~Mentzer, L.~V. Gool, and R.~Timofte, ``Learning for video compression with hierarchical quality and recurrent enhancement,'' in \emph{{Proc. IEEE Conf. Comp. Vis. Patt. Recogn.}}, 2020, pp. 6628--6637.

\bibitem{chen2021improving}
Z.~Chen, S.~Gu, F.~Zhu, J.~Xu, and R.~Zhao, ``Improving facial attribute recognition by group and graph learning,'' in \emph{Proc. IEEE Int. Conf. Multimedia Expo}.\hskip 1em plus 0.5em minus 0.4em\relax IEEE, 2021, pp. 1--6.

\bibitem{liu2024towards}
L.~Liu, Z.~Hu, and Z.~Chen, ``Towards point cloud compression for machine perception: A simple and strong baseline by learning the octree depth level predictor,'' \emph{arXiv preprint arXiv:2406.00791}, 2024.

\bibitem{li2021deep}
J.~Li, B.~Li, and Y.~Lu, ``Deep contextual video compression,'' \emph{{Advances in Neural Inf. Process. Syst.}}, vol.~34, pp. 18\,114--18\,125, 2021.

\bibitem{chen2022lsvc}
Z.~Chen, G.~Lu, Z.~Hu, S.~Liu, W.~Jiang, and D.~Xu, ``Lsvc: A learning-based stereo video compression framework. in 2022 ieee,'' in \emph{{Proc. IEEE Conf. Comp. Vis. Patt. Recogn.}}, 2022, pp. 6063--6072.

\bibitem{sheng2022temporal}
X.~Sheng, J.~Li, B.~Li, L.~Li, D.~Liu, and Y.~Lu, ``Temporal context mining for learned video compression,'' \emph{{{IEEE} Trans. Multimedia}}, 2022.

\bibitem{li2022hybrid}
J.~Li, B.~Li, and Y.~Lu, ``Hybrid spatial-temporal entropy modelling for neural video compression,'' in \emph{{Proc. {ACM} Int. Conf. Multimedia}}, 2022, pp. 1503--1511.

\bibitem{chen2023neural}
Z.~Chen, L.~Relic, R.~Azevedo, Y.~Zhang, M.~Gross, D.~Xu, L.~Zhou, and C.~Schroers, ``Neural video compression with spatio-temporal cross-covariance transformers,'' in \emph{{Proc. {ACM} Int. Conf. Multimedia}}, 2023, pp. 8543--8551.

\bibitem{chen2024group}
Z.~Chen, L.~Zhou, Z.~Hu, and D.~Xu, ``Group-aware parameter-efficient updating for content-adaptive neural video compression,'' in \emph{Proceedings of the 32nd ACM International Conference on Multimedia}, 2024, pp. 11\,022--11\,031.

\bibitem{duan2020video}
L.~Duan, J.~Liu, W.~Yang, T.~Huang, and W.~Gao, ``Video coding for machines: A paradigm of collaborative compression and intelligent analytics,'' \emph{{{IEEE} Trans. Image Process.}}, vol.~29, pp. 8680--8695, 2020.

\bibitem{pfeiffer2020adapterhub}
J.~Pfeiffer, A.~R{\"u}ckl{\'e}, C.~Poth, A.~Kamath, I.~Vuli{\'c}, S.~Ruder, K.~Cho, and I.~Gurevych, ``{A}dapter{H}ub: A framework for adapting transformers,'' in \emph{{Proc. Conf. Empirical Methods in Natural Language Processing}}, Oct. 2020, pp. 46--54.

\bibitem{he2021towards}
J.~He, C.~Zhou, X.~Ma, T.~Berg-Kirkpatrick, and G.~Neubig, ``Towards a unified view of parameter-efficient transfer learning,'' in \emph{{Proc. Int. Conf. Learn. Representations}}, 2022.

\bibitem{zhu2021counter}
Y.~Zhu, J.~Feng, C.~Zhao, M.~Wang, and L.~Li, ``Counter-interference adapter for multilingual machine translation,'' in \emph{{Proc. Conf. Empirical Methods in Natural Language Processing}}, 2021, pp. 2812--2823.

\bibitem{he2022sparseadapter}
S.~He, L.~Ding, D.~Dong, J.~Zhang, and D.~Tao, ``{S}parse{A}dapter: An easy approach for improving the parameter-efficiency of adapters,'' in \emph{{Proc. Conf. Empirical Methods in Natural Language Processing}}, Dec. 2022, pp. 2184--2190.

\bibitem{karimi2021compacter}
R.~Karimi~Mahabadi, J.~Henderson, and S.~Ruder, ``Compacter: Efficient low-rank hypercomplex adapter layers,'' \emph{{Advances in Neural Inf. Process. Syst.}}, vol.~34, pp. 1022--1035, 2021.

\bibitem{edalati2022krona}
A.~Edalati, M.~Tahaei, I.~Kobyzev, V.~P. Nia, J.~J. Clark, and M.~Rezagholizadeh, ``Krona: Parameter efficient tuning with kronecker adapter,'' \emph{Advances in Neural Inf. Process. Syst.}, 2022.

\bibitem{lee2022selective}
J.~Lee, S.~Jeong, and M.~Kim, ``Selective compression learning of latent representations for variable-rate image compression,'' \emph{{Advances in Neural Inf. Process. Syst.}}, vol.~35, pp. 13\,146--13\,157, 2022.

\bibitem{resnet}
K.~He, X.~Zhang, S.~Ren, and J.~Sun, ``Deep residual learning for image recognition,'' in \emph{{Proc. IEEE Conf. Comp. Vis. Patt. Recogn.}}, 2016, pp. 770--778.

\bibitem{zhao2018psanet}
H.~Zhao, Y.~Zhang, S.~Liu, J.~Shi, C.~C. Loy, D.~Lin, and J.~Jia, ``Psanet: Point-wise spatial attention network for scene parsing,'' in \emph{{Proc. Eur. Conf. Comp. Vis.}}, 2018, pp. 267--283.

\bibitem{frcnn}
S.~Ren, K.~He, R.~Girshick, and J.~Sun, ``Faster r-cnn: Towards real-time object detection with region proposal networks,'' \emph{{Advances in Neural Inf. Process. Syst.}}, vol.~28, 2015.

\bibitem{gumbel}
E.~Jang, S.~Gu, and B.~Poole, ``Categorical reparametrization with gumble-softmax,'' in \emph{{Proc. Int. Conf. Learn. Representations}}, 2017.

\bibitem{tsn}
L.~Wang, Y.~Xiong, Z.~Wang, Y.~Qiao, D.~Lin, X.~Tang, and L.~Van~Gool, ``Temporal segment networks: Towards good practices for deep action recognition,'' in \emph{{Proc. Eur. Conf. Comp. Vis.}}\hskip 1em plus 0.5em minus 0.4em\relax Springer, 2016, pp. 20--36.

\bibitem{xmem}
H.~K. Cheng and A.~G. Schwing, ``Xmem: Long-term video object segmentation with an atkinson-shiffrin memory model,'' in \emph{{Proc. Eur. Conf. Comp. Vis.}}\hskip 1em plus 0.5em minus 0.4em\relax Springer, 2022, pp. 640--658.

\bibitem{perazzi2016benchmark}
F.~Perazzi, J.~Pont-Tuset, B.~McWilliams, L.~Van~Gool, M.~Gross, and A.~Sorkine-Hornung, ``A benchmark dataset and evaluation methodology for video object segmentation,'' in \emph{{Proc. IEEE Conf. Comp. Vis. Patt. Recogn.}}, 2016, pp. 724--732.

\bibitem{kingma2015adam}
D.~P. Kingma and J.~Ba, ``Adam: A method for stochastic optimization,'' in \emph{{Proc. Int. Conf. Learn. Representations}}, 2015.

\bibitem{BPG}
B.~Fabrice, ``Bpg image format,'' 2018, \url{http://bellard.org/bpg/}, accessed:2022.

\end{thebibliography}

% \newpage

% \section{Biography Section}
% If you have an EPS/PDF photo (graphicx package needed), extra braces are
%  needed around the contents of the optional argument to biography to prevent
%  the LaTeX parser from getting confused when it sees the complicated
%  $\backslash${\tt{includegraphics}} command within an optional argument. (You can create
%  your own custom macro containing the $\backslash${\tt{includegraphics}} command to make things
%  simpler here.)
 
% \vspace{11pt}

% \bf{If you include a photo:}\vspace{-33pt}
% \begin{IEEEbiography}[{\includegraphics[width=1in,height=1.25in,clip,keepaspectratio]{fig1}}]{Michael Shell}
% Use $\backslash${\tt{begin\{IEEEbiography\}}} and then for the 1st argument use $\backslash${\tt{includegraphics}} to declare and link the author photo.
% Use the author name as the 3rd argument followed by the biography text.
% \end{IEEEbiography}

% \vspace{11pt}

% \bf{If you will not include a photo:}\vspace{-33pt}
% \begin{IEEEbiographynophoto}{John Doe}
% Use $\backslash${\tt{begin\{IEEEbiographynophoto\}}} and the author name as the argument followed by the biography text.
% \end{IEEEbiographynophoto}

% \vfill

\end{document}